\documentclass[letterpaper]{elsarticle} 

\usepackage{graphics}
\usepackage{comment}
\usepackage{bmpsize}
\usepackage{amsmath}
\usepackage{amsthm}
\usepackage{bm}
\usepackage{amssymb}
\usepackage{mathtools}
\usepackage{setspace}
\usepackage{enumitem}
\usepackage{pstricks}
\usepackage{epstopdf}
\usepackage[linesnumbered,ruled,vlined]{algorithm2e}
\usepackage{microtype}
\usepackage{xcolor}
\usepackage{multirow}

\theoremstyle{plain}            

\theoremstyle{definition}   
\newtheorem{defn}{\protect\definitionname}
\theoremstyle{definition}

\theoremstyle{definition}

\providecommand{\definitionname}{Definition} 
\providecommand{\problemname}{Problem Definition} 
\providecommand{\extensionname}{Extended Problem Definition} 

\newcommand{\rl}{\textsc{RL}}
\newcommand{\irl}{\textsc{IRL}}
\newcommand{\mdp}{\textsc{MDP}}
\newcommand{\pomdp}{\textsc{POMDP}}
\newcommand{\M}{\mathcal{M}}
\newcommand{\highlight}[1]{\textcolor{blue}{#1}}

\begin{document}
	
	\begin{frontmatter} 
	
          \title{A   Survey   of   Inverse   Reinforcement   Learning:
            Challenges, Methods and Progress}
		
          \author{Saurabh Arora} \address{THINC Lab, Dept. of Computer Science, University of Georgia, Athens, GA 30602} \ead{sa08751@uga.edu}     
          \author{Prashant     Doshi}
          \address{THINC Lab and Institute for AI, Dept. of Computer Science, University of
            Georgia, Athens, GA 30602} \ead{pdoshi@cs.uga.edu}

          \begin{abstract}
            Inverse reinforcement learning (\irl{})  is the problem of
            inferring  the  reward function  of  an  agent, given  its
            policy or  observed behavior.  Analogous to  \rl{}, \irl{}
            is perceived both as a problem  and as a class of methods.
            By  categorically  surveying  the  current  literature  in
            \irl{}, this article serves as a reference for researchers
            and  practitioners  of  machine  learning  and  beyond  to
            understand  the  challenges  of   \irl{}  and  select  the
            approaches  best  suited for  the  problem  on hand.   The
            survey formally  introduces the \irl{} problem  along with
            its  central   challenges  such   as  the   difficulty  in
            performing  accurate inference  and its  generalizability,
            its    sensitivity   to    prior   knowledge,    and   the
            disproportionate  growth   in  solution   complexity  with
            problem  size.  The  article  elaborates  how the  current
            methods mitigate these challenges.  We further discuss the
            extensions  to traditional  \irl{}  methods for  handling:
            inaccurate and incomplete perception, an incomplete model,
            multiple reward functions, and nonlinear reward functions.
            This  survey  concludes  the discussion  with  some  broad
            advances in the research  area and currently open research
            questions.
          \end{abstract}
		
		
	\end{frontmatter}

	\singlespacing

\section{Introduction}
\label{sec:intro}
	
Inverse reinforcement learning (\irl{}) is the problem of modeling the
preferences  of another  agent  using its  observed behavior,  thereby
avoiding     a     manual      specification     of     its     reward
function~\citep{Russell1998,Ng2000}.  In the past decade or so, \irl{}
has  attracted several  researchers in  the communities  of artificial
intelligence, psychology, control theory, and machine
learning. 
IRL is  appealing because  of its  potential to  use data  recorded in
performing  a task  to  build autonomous  agents  capable of  modeling
others without intervening in the performance of the task.

\begin{figure}[ht!]
  \centering{
  	\includegraphics{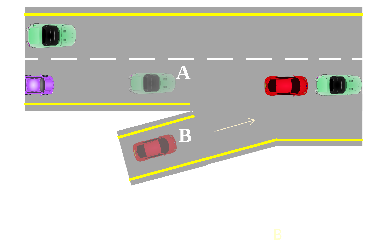}
        \caption{Red car B is trying to merge into the lane, and green
          car A is  the immediate traffic.  The  transparent images of
          cars  show their  positions before  merging, and  the opaque
          images to the  right depict one of  their possible positions
          after the merger.}
    \label{fig:car_merger}
  }
\end{figure}

We study this  problem and associated advances in a  structured way to
address the needs of readers with different levels of familiarity with
the field.  For  clarity, we use a contemporary  example to illustrate
\irl{}'s use  and associated challenges.  Consider  a self-driving car
in  role  B in  Fig.~\ref{fig:car_merger}.   To  safely merge  into  a
congested freeway,  it may model  the behavior of  the car in  role A;
this car forms the immediate  traffic. We may use previously collected
trajectories of  cars in  roles A  and B, on  freeway entry  ramps, to
learn the  safety and  speed preferences  of a  typical driver  as she
approaches  this  difficult  merge  (NGSIM~\cite{NGSIM}  is  one  such
existing data set).

Approaches for \irl{} predominantly  ascribe a Markov decision process
(\mdp{})~\cite{Puterman1994} to the interaction  of the observed agent
with its environment,  and whose solution is a {\em  policy} that maps
states to actions. The reward function  of this \mdp{} is unknown, and
the  observed agent  is assumed  to follow  an optimal  policy of  the
\mdp{}.   In the  traffic  merge example,  the  \mdp{} represents  the
driving process  of car A.   The driver of  Car A is  following action
choices (deceleration, braking, low acceleration, and others) based on
its optimal policy.  Car B needs to  reach the end of the merging lane
after Car A for merging safely.
	

\subsection{Significance of \irl{}}
\label{subsec:significance}

Researchers  in   the  areas   of  machine  learning   and  artificial
intelligence have  developed a substantial interest  in \irl{} because
it caters to the following needs.
	
\subsubsection{Demonstration substitutes manual specification
  of reward}
\label{subsubsection:apprenticeship}

Typically, if a  designer wants intelligent behavior in  an agent, she
manually  formulates the  problem  as a  forward  learning or  forward
control  task solvable  using  solution techniques  in \rl{},  optimal
control, or predictive control. A key element of this formulation is a
specification  of  the agent's  preferences  and  goals via  a  reward
function.  In the  traffic merge example, we may hand  design a reward
function for Car  A. For example, +1  reward if taking an  action in a
state decreases the relative velocity of  Car B w.r.t.  Car A within a
predefined distance from merging junction, thereby allowing for a safe
merge. Analogously, a  negative reward of -1 if taking  an action in a
state increases the relative velocity of Car B w.r.t. Car A.
This example  specification captures one aspect  of a
  successful merge into a congested freeway: that the merging car must
  slow down to match the speed  of the freeway traffic. However, other
  aspects such  as merging a  safe distance behind  Car A and  not too
  close in  front of the  car behind A  require further tuning  of the
  reward  function.   While  roughly specified  reward  functions  are
  sufficient  in  many domains  to  obtain  expected behavior  (indeed
  affine transformations of the  true reward function are sufficient),
  others  require  much  trial-and-error  or  a  delicate  balance  of
  multiple  conflicting  attributes~\cite{Coates2009},  which  becomes
  cumbersome.

  The need to pre-specify the reward function limits the applicability
  of \rl{} and optimal control to problems where a reward function can
  be  easily   specified.   \irl{}  offers   a  way  to   broaden  the
  applicability  of   RL  and  reduce   the  manual  design   of  task
  specification, given  a policy or demonstration  of desired behavior
  is  available.   While  acquiring  the complete  desired  policy  is
  usually  infeasible,  we have  easier  access  to demonstrations  of
  behaviors, often in  the form of recorded data.   For example, state
  to action mappings for all contingencies for Car B are not typically
  available, but datasets such as NGSIM contain trajectories of Cars A
  and B  in real-world driving.  Thus,  \irl{} forms a key  method for
  \emph{learning from demonstration}~\cite{argall2009survey}.
  
	
	
  A  topic in  control theory  related  to \irl{}  is inverse  optimal
  control~\cite{Boyd94}.  While  the input in both  \irl{} and inverse
  optimal control  are trajectories consisting of  state-action pairs,
  the  target of  learning in  inverse optimal  control is  a function
  mapping states of observed agent  to her actions. The learning agent
  may use this policy  to imitate it or deliberate with  it in its own
  decision-making process.

		
\subsubsection{Improved Generalization} 

A reward function represents the preferences of an agent in a succinct
form, and is amenable to transfer to another agent. The learned reward
function  may be  used as  is  if the  subject agent  shares the  same
environment and goals as the  other, otherwise it continues to provide
a  useful  basis when  the  agent  specifications differ  mildly,  for
example, when  the subject agent's problem  domain exhibits additional
states.
Indeed, as Russell~\cite{Russell1998} points out, the
  reward function  is inherently  more {transferable} compared  to the
  observed agent's policy. This is  because even slight changes in the
  environment  -- for  example, changes  to  the noise  levels in  the
  transition function  -- likely  renders the learned  policy unusable
  because  it may  not be  directly revised  in straightforward  ways.
  However,  this change  does not  impact the  transferability of  the
  reward function. Furthermore,  it is likely that  the learned reward
  function  simply needs  to  be extended  to any  new  states in  the
  learner's environment while  a learned policy would  be discarded if
  the new states are significant.


\subsubsection{Potential Applications} 

While  introducing \irl{},  Russell~\cite{Russell1998} alluded  to its
potential application  in providing computational models  of human and
animal  behavior because  these  are difficult  to  specify.  In  this
regard,   Baker   et   al.~\cite{Baker2009_action}   and   Ullman   et
al.~\cite{Ullman2009} demonstrate  the inference of a  human's goal as
an inverse planning  problem in an \mdp{}.   Furthermore, \irl{}'s use
toward apprenticeship learning has rapidly expanded the set of visible
use cases. These can be categorized into:

  \begin{figure}[ht!]
    \centering{
    \includegraphics[clip=true,trim=0             0.5cm            0      0.005cm]{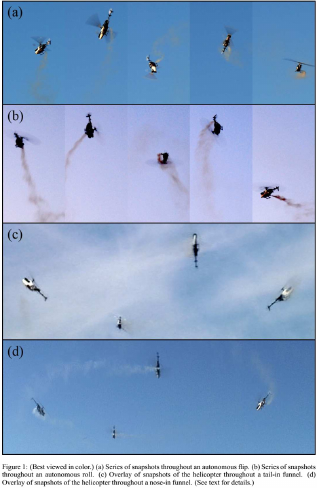}
    \caption{Complex  helicopter maneuvers  learned using  \rl{} on  a
      reward  function learned  from an  expert pilot  through \irl{}.
      The image  is reprinted  from \cite{Abbeel2007}  with permission
      from publisher.}\label{HelicopterManeuver}
    }
  \end{figure}
  
\begin{enumerate}[itemsep=0in]

\item Learning  from an expert  to create  an agent with  the expert's
  preferences.   An  early  and well-known  application  that  brought
  significant    attention   to    \irl{}    is   helicopter    flight
  control~\cite{Abbeel2007},               illustrated              in
  Fig.~\ref{HelicopterManeuver}.   In  this   application,  an  expert
  helicopter  operator's sophisticated  preferences  over 24  features
  were learned from  recorded behavior data using  \irl{}. This reward
  function  was  then used  to  teach  a physical  remotely-controlled
  helicopter advanced maneuvers using  \rl{}. Another application that
  brought \irl{} closer  to Russell's~\cite{Russell1998} motivation of
  modeling animal behavior is that  of socially adaptive navigation to
  avoid colliding into humans by learning from human-walk trajectories
  \cite{Kretzschmar_interactingpedestrians,Kim2016_adaptivenavigation}.
  Other  important examples  include boat  sailing~\cite{Neu2007}, and
  learning driving styles~\cite{Kuderer15:Learning}.

\item   Learning  from   another  agent   to  predict   its  behavior.
  One  of the  first attempts  in this  direction was
    route  prediction   for  taxis~\cite{Ziebart2008,Ziebart_Cabbie}.
  Other such applications are  footstep prediction for planning legged
  locomotion~\cite{Ratliff2009},     anticipation    of     pedestrian
  interactions~\cite{Ziebart_predictionpedestrian},  energy  efficient
  driving~\cite{Vogel_efficientdriving},   and    penetration   of   a
  perimeter  patrol  by  learning   the  patrollers'  preferences  and
  patrolling route~\cite{Bogert_mIRL_Int_2014}.
\end{enumerate}

\subsection{Importance of this Survey}
\label{subsec:target_survey}
This article  is a reflection  on the research  area of \irl{}  with a
focus on the following important aspects:
	
\begin{enumerate}[topsep=0in,itemsep=0in]
\item  Formally introducing  \irl{} and  its importance,  by means  of
  various examples,  to the researchers  and practitioners new  to the
  field;
\item A  study of the  challenges that make \irl{}  difficult,
  and a review of the current (partial) solutions;
\item Qualitative assessment and  comparisons among different methods,
  both those that  are foundational and those that  are extensions, to
  evaluate them  coherently. This  will considerably help  the readers
  decide on the approach suitable for the problem at hand;
\item Identification  of some  significant milestones achieved  by the
  methods in this field;
\item Identification of the common shortcomings and open avenues for future research.
\end{enumerate}
	
Of course, a single article may  not cover all methods in this growing
field.  Nevertheless,   we  have  sought   to  make  this   survey  as
comprehensive as possible.
	 
\subsection{Organization of Contents}
\label{subsec:organization}


As \irl{}  is relatively  new and  the target reader  is likely  to be
someone who is keen to learn about  \irl{}, the viewpoint of `IRL as a
research problem' is  used to guide the organization  of this article.
Therefore,  Section~\ref{sec:IRL}  mathematically defines  the  \irl{}
problem  and provides  some preliminary  technical background  that is
referenced in later  sections. We introduce the  core challenges faced
by  this  learning  problem  in  Section~\ref{sec:challenges}.   These
challenges confront all methods and are not specific to any particular
technique.  Then,  we briefly review the  foundational methods grouped
together  by   the  commonality   of  their  underlying   approach  in
Section~\ref{sec:PrimaryMethods}  that  have facilitated  progress  in
\irl{}, and how these  methods mitigate the previously-introduced core
challenges in Section~\ref{sec:Mitigation}.  This separation of method
description  across  two sections  allows  a  practitioner to  quickly
identify the methods pertinent to  the most egregious challenge she is
facing    in   her    \irl{}   problem.     This   is    followed   in
Section~\ref{sec:Extensions} by a review of efforts that generalize or
extend the fundamental \irl{} problem in various directions.
Finally, the article concludes with  a discussion of some shortcomings
and open research questions.


\section{Formal Definition of \irl{}}
\label{sec:IRL}

In  order  to  formally define  \irl{},  we  must  first decide  on  a
framework  for modeling the  observed agent's  behavior. While
methods    ascribe   different frameworks   such   as   an   \mdp{},
hidden-parameter \mdp{}, or a \pomdp{}  to the expert, we focus on the
most popular model by far, which is the \mdp{}. 

	
\begin{defn}\label{definition:mdp}[\mdp{}]
  An \mdp{}  $\M{}:=\langle S,A,T,R,\gamma \rangle$ models  an agent's
  sequential decision-making  process. $S$ is  a finite set  of states
  and     $A$      is     a      set     of      actions.      Mapping
  $T:S\times   A   \to   \textsf{Prob}(S)$   defines   a   probability
  distribution over  the set of  next states conditioned on  the agent
  taking action $a$ at state  $s$; $\textsf{Prob}(S)$ here denotes the
  set of all probability  distributions over $S$.  $T(s'|s,a)\in[0,1]$
  is the probability that the system transitions to state $s'$.
  The  reward  function  $R$   can  be  specified  in
    different   ways:   $R:S   \to  \mathbb{R}$   gives   the   scalar
    reinforcement at state  $s$, $R:S \times A \to  \mathbb{R}$ maps a
    tuple (state  $s$, action $a$  taken in  state $s$) to  the reward
    received      on       performing      the       action,      and,
    $R:S \times A \times S \to  \mathbb{R}$ maps a triplet (state $s$,
    action  $a$,  resultant state  $s'$)  to  the reward  obtained  on
    performing the transition.
    Discount factor $\gamma \in [0,1]$  is the weight for past rewards
    accumulated              in             a              trajectory,
    $\langle                       (s_{0},a_{0}),(s_{1},a_{1}),\ldots$
    $(s_{j},a_{j})                   \rangle$,                   where
    $s_{j}\in S,a_{j}\in A,j \in\mathbb{N}$.
\end{defn}
	
A \emph{policy}  is a  function mapping current  state to  next action
choice(s).  It  can  be  deterministic,  $\pi:S\to  A$  or  stochastic
$\pi:S\to  \textsf{Prob}(A)$. 
For  a  policy $\pi$,  value  function
$V^{\pi}:S  \to \mathbb{R}$  gives the  value of  a state  $s$  as the
long-term  expected  cumulative  reward  incurred from  the  state  by
following $\pi$. The value of a policy $\pi$ for a given start state
$s_0$ is,

\begin{equation}
  V^{\pi}(s_0) = E_{s,\pi(s)}\left[ \sum_{t=0}^\infty \gamma^t
    R(s_t,\pi(s_t))|s_0 \right]
\label{eqn:state_value}
\end{equation}

The goal of solving \mdp{} $\M{}$ is to find an optimal policy $\pi^*$
such that  $V^{\pi^*}(s)=V^*(s)=\sup_\pi~ V^{\pi}(s)$, for
all    $s\in   S$.     The    action-value    function   for    $\pi$,
$Q^{\pi}:S\times A  \to \mathbb{R}$, maps  a state-action pair  to the
long-term expected cumulative reward  incurred after taking action $a$
from $s$  and following policy  $\pi$ thereafter.  We also  define the
optimal  action-value  function as  $Q^*(s,a)=\sup_\pi  Q^{\pi}(s,a)$.
Subsequently, $V^*(s) = \sup_{a \in A} Q^*(s,a)$.  Another perspective
to  the  value  function  involves multiplying  the  reward  with  the
converged {\em state-visitation frequency} $ \psi^{\pi}(s) $, which is
the number  of times  the state  s is  visited on  using policy  $ \pi
$. The latter is given by:
\begin{align}
\psi^{\pi}(s) = \psi^0(s) + \gamma \sum_{s' \in S} T(s,\pi(s),s')~\psi^{\pi}(s')
\label{eqn:visit_freq}
\end{align}
where $ \psi^0(s) $ is initialized as  0 for all states.  Let $ \Psi $
be   the   space    of   all   $   \psi    $   functions.    Iterating
Eq.~\ref{eqn:visit_freq}  until the  state-visitation frequency  stops
changing yields the converged  frequency function, $\psi^{\pi}_*$.  We
may         write        the         value        function         as,
$V^*(s)      =      \underset{\pi}{\sup}~     \sum_{s      \in      S}
\psi^{\pi}_*(s)~R(s,\pi(s))$.

We  may express  the  reward  function as  a  linear  sum of  weighted
features:
\begin{align}
R(s,a) & = w_1 \phi_1(s,a) + w_2 \phi_2(s,a) + \ldots + w_k \phi_k(s,a)\nonumber\\ 
& = \bm{w}^T\bm{\phi}(s,a).
\label{eqn:linear-reward}  
\end{align}
where  $\phi_k:S\to  \mathbb{R}$  is  a feature  function  and  weight
$w_k \in  \mathbb{R}$. Then, define  the expected feature count for
policy $\pi$ and feature $\phi_k$ as,
\begin{align}
  \mu^{\phi_k}(\pi)= \sum_{t=0}^\infty \psi^\pi(s_t)~\phi_k(s_t,\pi(s_t)).  
\label{eqn:feature-expectation}
\end{align}
We will extensively  refer to this formulation of  the reward function
and  the expected  feature  count  later in  this  article. Note  that
$\mu^{\phi_k}(\pi)$ is also called a successor feature in RL. The
expected feature count can be used to define the expected value of a policy:
\begin{align}
  V^\pi &= \bm{w}^T\bm{\mu}^{\mathbf{\phi}}(\pi) = \sum_{s,a} \psi^{\pi}(s)~\bm{w}^T \bm{\phi}(s,a)\nonumber\\
        &=\sum_{s,a} \psi^\pi(s)~R(s,a).
\label{eqn:value-function-feature}
\end{align}
		


\rl{} offers  an online way  to solve an  \mdp{}. The
  input for \rl{}  is the sequence of sampled experiences  in the form
  $(s,  a, r)$  or  $(s, a,  r,  s')$, which  includes  the reward  or
  reinforcement due to  the agent performing action $a$  in state $s$.
  For the model-free setting of  \rl{}, the transition function $T$ is
  unknown. Both the transition function  and policy are estimated from
  the samples and the target of \rl{} is to learn an optimal policy.
	
We  adopt the  conventional terminology  in \irl{},  referring to  the
observed agent  as an {\em expert}  and the subject agent  as the {\em
  learner}. 
Typically, \irl{} assumes that the expert is behaving
  according  to  an  underlying  policy  $\pi_E$,  which  may  not  be
  known. If policy is not known, the learner observes sequences of the
  expert's  state-action   pairs  called  trajectories.    The  reward
  function is unknown  but the learner usually  assumes some structure
  that  helps in  the  learning.  Common  functional  forms include  a
  linearly-weighted  combination  of  reward features,  a  probability
  distribution   over   reward   functions,  or   a   neural   network
  representation.  We elaborate on these  forms later in this article.
  The expert's transition  function may not be known  to the learner.
We are now ready to give the formal problem definition of \irl{}.
 
	
\begin{defn}[\irl{}]     Let     an     \mdp{}     without     reward,
  $\M\backslash_{R_E}$, model  the interaction of the  expert $E$ with
  the environment.
   Let                                           $\mathcal{D}=\{\langle
   (s_{0},a_{0}),(s_{1},a_{1}),\ldots,
   (s_{j},a_{j})   \rangle_1,$ $\ldots, \langle
   (s_{0},a_{0}),(s_{1},a_{1}),$ $\ldots,
   (s_{j},a_{j})   \rangle_{i=2}^{N}   \}$, $s_{j}\in  S$, $a_{j}\in
   A$, and $i,j,N \in\mathbb{N}$ 
   be  the   set  of  demonstrated  trajectories.    A  trajectory  in
   $\mathcal{D}$  is  denoted  as  $\tau$.  We  may  assume  that  all
   $\tau  \in \mathcal{D}$  are  perfectly  observed. Then,  determine
   $\hat{R}_E$ that  best explains either  policy $\pi_E$ if  given or
   the observed behavior in the form of demonstrated trajectories.
  \label{def:irl}
\end{defn}

\begin{figure}[ht!]
  \centering{
      \includegraphics[height=2.25in,width=5.0in]{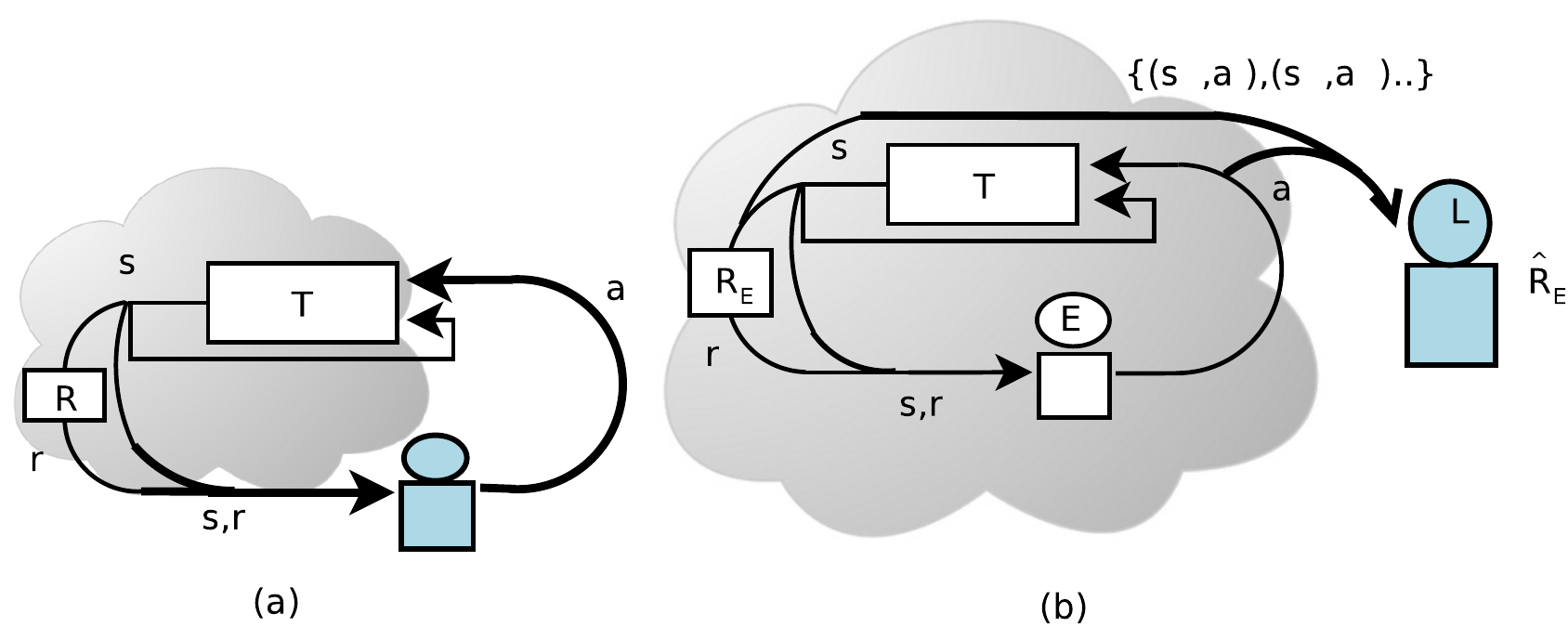}
      \caption{$(a)$ A schematic showing  the subject agent (shaded in
        blue)   performing  \rl{}~\cite{Kaelbling1996}.    In  forward
        learning or RL,  the agent chooses an action at  a known state
        and receives a reward in return generated by a reward function
        $R$ that  may not  be known  to the  agent. The  state changes
        based on the previous state and action, which is modeled using
        the  transition function  $T$  that may  be  unknown as  well.
        $(b)$ In inverse learning or  \irl{}, the input and output for
        the learner $L$ are reversed.  
        L       perceives      the       states      and       actions
        $\{(s,a),(s,a),\ldots,  (s,a)\}$ of  expert E  (or its  policy
        $\pi_E$), and  learns a reward function  $\hat{R}_E$ that best
        explains $E$'s behavior, as the  output. Note that the learned
        reward function may not exactly  correspond to the true reward
        function. 
    }
    \label{fig:IRL}
  }
\end{figure}

Notice that \irl{} inverts the \rl{} problem. %
Whereas  \rl{} seeks  to learn  the optimal  behavior
  based on experiences  ($(s, a, r)$ or $(s, a,  r, s')$) that include
  obtained rewards, \irl{} seeks to best explain the observed behavior
  by learning the corresponding  reward function.  We illustrate this
relationship between \rl{} and \irl{} in
Fig.~\ref{fig:IRL} 

We may obtain  an estimate of the expected feature  count from a given
demonstration  $\mathcal{D}$   of  $N$  trajectories,  which   is  the
empirical analog of that in Eq.~\ref{eqn:feature-expectation},
\begin{equation}
\label{eqn:empirical-feature-expectation}
\hat{\mu}^{\phi_k}(\mathcal{D})=\frac{1}{N} \sum_{i=1}^N\sum_{t=0}^\infty \gamma^t
\phi_k(s_t,a_t).
\end{equation}

\section{Primary Challenges of \irl{}}
\label{sec:challenges} 

\irl{} is challenging because the optimization associated in finding a
reward  function  that   best  explains  observations  is  essentially
ill-posed.  Furthermore,  computational costs of  solving
 the problem
tend to grow disproportionately with the size of the problem. We
discuss  these   challenges  in  detail  below,  but   prior  to  this
discussion,  we establish  some notation.  Let $\hat{\pi}_E$  be the
policy obtained by  optimally solving the \mdp{}  with  reward 
function $\hat{R}_E$. 


\subsection{Obstacles to Accurate Inference} 
\label{subsection:accuracy}

\begin{figure}[ht!]
    	\centering
    	\includegraphics[scale=.4]{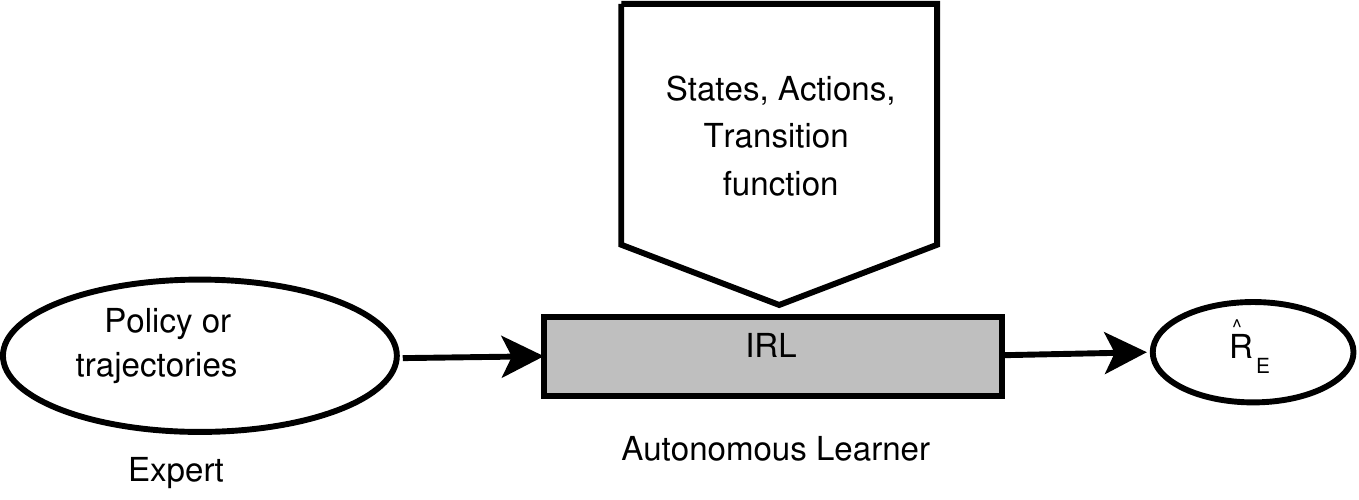} 
    	\caption{Pipeline for a classical  \irl{} process. The learner
          receives an  optimal policy  or trajectories as  input.  The
          prior domain  knowledge (shown  here as a  pentagon) include
          the  completely observable  state space,  action space,  and
          fully known transition probabilities.}
   	\label{fig:ideal_IRL}
\end{figure}

Classical \irl{} takes an expert demonstration of a task consisting of
a  finite  set  of  trajectories, knowledge  of  the  environment  and
expert's dynamics,  and finds the expert's  potential reward function;
this is illustrated in Fig.~\ref{fig:ideal_IRL}.

A critical  challenge, first noticed by  Ng and Russell~\cite{Ng2000},
is that many  reward functions (including highly  degenerate ones such
as  a  function  with  all  reward  values  zero)  could  explain  the
observations.  This is because the input is usually a finite and small
set of trajectories (or a policy) and many reward functions in the set
of  all  reward  functions  can generate  policies  that  realize  the
observed  demonstration. Thus,  \irl{}  suffers from  an ambiguity  in
solution.

Given the difficulty of ensuring  accurate inference, its pertinent to
contemplate how we  may measure accuracy. If the  true reward function
$R_E$ is  available for  purposes of evaluation,  then one  measure of
accuracy is the closeness of  a learned reward function $\hat{R}_E$ to
$R_E$, $\left|\left|R_E-\hat{R}_E\right|\right|_p$.  However, a direct
comparison of rewards is not useful because an \mdp{}'s optimal policy
is   invariant   under   affine    transformations   of   the   reward
function~\cite{Russell03:Artificial}.  On  the other hand,  two reward
functions  similar   for  the  most   part  but  differing   for  some
state-action  pairs may  produce considerably  different policies  and
behaviors.   To make  the  evaluation targeted,  a  comparison of  the
behavior  generated from  the learned  reward function  with the  true
behavior of expert is more appropriate. In other words, we may compare
the policy  $\hat{\pi}_E$ generated from \mdp{}  with $\hat{R}_E$ with
the true  policy $\pi_E$. The  latter could  be given or  is generated
using  the true  reward function.   A  limitation of  this measure  of
accuracy is  that a difference  between the  two policies in  just one
state  could  still  have  a  significant  impact.   This  is  because
performing  the  correct  action  at  that state  may  be  crucial  to
realizing  the  task.   Consequently,  this measure  of  closeness  is
inadequate because it would report just a small difference despite the
high significance.

This brings us to the conclusion that we should measure the difference
in  values of  the learned  and  true policies.  Specifically, we  may
measure the  error in inverse  learning, called  {\em inverse
  learning    error}     (ILE),    as    $\left|\left|V^{\pi_E}    -
    V^{\hat{\pi}_E}          \right|\right|_p$
where $V^{\pi_E}$  is the value function  for actual policy $  \pi_E $
and  $V^{\hat{\pi}_E}$ is  that for  the learned  policy $\hat{\pi}_E$
both obtained using the true reward function~\cite{Choi2011}.
Notice that if the true and learned policies are the same, then ILE is
zero. However,  ILE may also  vanish when the  two differ but  if both
policies are  optimal.  On  the other hand,  ILE requires  knowing the
true reward  function which limits  its use to  formative evaluations.
Another assessment measures the {\em learned behavior accuracy}.  This
metric is  computed as the  number of demonstrated  state-action pairs
that match between using the true  and learned policies expressed as a
percentage of the former.

\subsection{Generalizability}
\label{subsection:generalization_capacity} 

Generalization refers  to the extrapolation of  learned information to
the states and actions unobserved in the demonstration and to starting
the task at different initial states.  Observed trajectories typically
encompass a subset  of the state space and the  actions performed from
those
states. 
Well-generalized  reward  functions  should reflect  expert's  overall
preferences  relevant to  the  task. The  challenge  is to  generalize
correctly  to the  unobserved space  using  data that  often covers  a
fraction of the complete space.

Notice  that achieving  generalizability  promotes  the temptation  of
training  the learner  using  fewer examples  because  the latter  now
possesses  the   ability  to  extrapolate.  However,   less  data  may
contribute  to  greater  approximation  error in  $  \hat{R}_E  $  and
inaccurate inference.

ILE  continues  to be  pertinent  by offering  a  way  to measure  the
generalizability of  the learned information as well.  This is because
it   compares   value   functions,   which  are   defined   over   all
states. Another procedure for evaluating generalizability is to simply
withhold   a   few  of   the   demonstration  trajectories  from   the
learner. These can be used as labeled test data for comparing with the
output of the learned policy on the undemonstrated state-action pairs.


\subsection{Sensitivity to Correctness of Prior Knowledge}
\label{subsection:sensitivity_priorknowledge}
If we represent the reward  function, $R_E$, as a weighted combination
of feature functions,  the problem then reduces to  finding the values
of       the       weights.         Each       feature       function,
$\phi: S \times A \to \mathbb{R}$, is given and is intended to model a
facet of the expert's preferences.

Prior  knowledge  enters  \irl{}  via  the  specification  of  feature
functions in $R_E$ and the  transition function in the \mdp{} ascribed
to the  expert. Consequently, the  accuracy of \irl{} is  sensitive to
the selection of feature functions that not only encompass the various
facets of the expert's true reward function but also differentiate the
facets.   Indeed,  Neu  and Szepesv\'{a}ri~\cite{Neu2007}  prove  that
\irl{}'s  accuracy  is   closely  tied  to  the   scaling  of  correct
features. Furthermore, it is also  dependent on how accurately are the
dynamics of the expert modeled by the ascribed \mdp{}. If the dynamics
are  not  deterministic,  due  to  say  some  noise  in  the  expert's
actuators,  the  corresponding  stochasticity needs  to  be  precisely
modeled in the transitions.

Given the significant role of prior knowledge in \irl{}, the challenge
is two-fold:  $(i)$ we  must ensure  its accuracy,  but this  is often
difficult  to  achieve   in  practice;  $(ii)$  we   must  reduce  the
sensitivity of solution methods to  the correctness of prior knowledge
or replace the knowledge with learned information.

	
	
\subsection{Disproportionate  Growth   in  Solution   Complexity  with
  Problem Size}
\label{subsection:labelCompExpense} 

Methods for \irl{} are iterative  as they involve a constrained search
through the space of reward functions.  As the  number of iterations  may vary  based on
whether  the  optimization is  convex, it is  linear,  the  gradient can  be
computed quickly, or  none of these, we focus on analyzing the complexity of
each iteration.  Consequently, the computational
complexity is expressed as the time complexity of each iteration and its
space complexity.

Each iteration's  time is dominated  by the complexity of  solving the
ascribed \mdp{}  using the  reward function currently  learned.  While
the complexity of  solving an \mdp{} is polynomial in  the size of its
parameters, the parameters such as the state space are impacted by the
curse of  dimensionality -- its size  is exponential in the  number of
components of state vector  (dimensions). Furthermore, the state space
in  domains such  as robotics  is  often continuous  and an  effective
discretization also  leads to an  exponential blowup in the  number of
discrete states. Therefore, increasing  problem size adversely impacts
the run time of each iteration of \irl{} methods.

Another   type  of   complexity  affecting   \irl{}  is   {\em  sample
  complexity}, which refers  to the number of  trajectories present in
the input
demonstration. 
As  the  problem size  increases,  the  expert must  demonstrate  more
trajectories in  order to maintain  the required level of  coverage in
the training data.
\subsection{Direct Learning of Reward or Policy Matching}
\label{subsection:labelDirectVsIndirect} 

Two distinct  approaches to \irl{}  present themselves, each  with its
own attendant set  of challenges. First seeks  to directly approximate
the reward  function $\hat{R}_E$ by  tuning it using input  data.  The
second approach focuses on learning  a policy that matches its actions
or action values with the  demonstrated behavior, thereby learning the
reward function as an intermediate step.

Success  of the  first approach  hinges on  selecting an  adequate and
complete reward structure (for example,  the set of feature functions)
that composes the reward function.   Though learning a reward function
offers a  deeper generalization of  the task at  hand, it may  lead to
policies that do not fully reproduce the observed
trajectories. 
For the second approach, Neu and Szespesv\'{a}ri.~\cite{Neu2008} point
out that  the optimization  for \irl{}  is convex  if the  actions are
deterministic and  the demonstration  spans the complete  state space.
While  both  approaches  are  negatively  impacted  by  reduced  data,
matching  the observed  policy  is particularly  sensitive to  missing
states in  the demonstration, which  makes the problem  non-convex and
weakens the objective of matching the given (but now partial) policy.


\vspace{0.1in} Next  section categorizes  the foundational  methods in
\irl{} based on the mathematical  framework they use for learning, and
discusses them in some detail.

\section{Foundational Methods for \irl{}}
\label{sec:PrimaryMethods} 

Many \irl{} methods fit a template of key steps. We show this template
in Algorithm~\ref{alg:IRL_template},  and present  the methods  in the
context of this  template. Such presentation allows us  to compare and
contrast various
methods. 
Algorithm~\ref{alg:IRL_template}  assumes   that  the
  expert's MDP sans the reward function  is known to the learner as is
  commonly  assumed  in  most  IRL  methods  although  a  few  methods
  discussed later allow the transition function to be unknown.  Either
  a demonstration or the expert's policy  is provided as input as well
  as any features for the reward function.

\begin{algorithm}[ht!]
	\caption{Template for \irl{}}	
	\label{alg:IRL_template}
	\KwIn{$\M\backslash_{R_E}$= $ \langle S,A,T,\gamma \rangle $,\\
          Set   of   trajectories  demonstrating   desired   behavior:
          $\mathcal{D}=\{\langle
          (s_0,a_0),(s_1,a_1),\ldots,(s_t,a_t)  \rangle, \ldots  \}$,  $s_t\in S$,
          $a_t\in A$,  $t \in \mathbb{N}$, \\
          or expert's policy: $\pi_E$, and
          reward function features}
        \KwOut{  $\hat{R}_E$} 
          Model the expert's observed  behavior as the solution of an \mdp{} whose
          reward  function is not known\; 

          Initialize  the parameterized  form of  the reward  function
          using any  given features (linearly weighted  sum of feature
          values, distribution over rewards, or other)\;

          Solve  the \mdp{} with  current reward function  to generate
          the learned behavior or policy\; 

          Update  the optimization parameters  to minimize  the
          divergence  between the  observed behavior (or policy)  and the  learned
          behavior (policy)\; 

          Repeat the previous two steps till the divergence
          is reduced to a desired level. 
\end{algorithm}

Existing  methods seek  to learn  the expert's  preferences, a  reward
function $\hat{R}_E$, represented in different  forms such as a linear
combination of weighted feature  functions, a probability distribution
over  multiple real-valued  maps  from states  to  reward values,  and
others.    Parameters   of  $\hat{R}_E$   vary   with   the  type   of
representation   (weights,   parameters    defining   the   shape   of
distribution).  \irl{}  involves solving the \mdp{}  with the function
hypothesized  in  current  iteration   and  updating  the  parameters,
constituting a search  that terminates when the  behavior derived from
the current solution aligns with the observed behavior.

Rest  of this  section categorizes  \irl{} methods  based on  the core
approach they use  for inverse learning --  margin based optimization,
entropy  based optimization,  Bayesian inference,  classification, and
regression. A  second-level grouping  within each of  these categories
clusters methods based on the  specific objective function utilized in
realizing  the  core  approach.   Recall the  notation  introduced  in
Section~\ref{sec:IRL} before continuing.

\subsection{Margin Optimization}
\label{subsec:max-margin}

Maximum  margin  prediction  aims  to learn  a  reward  function  that
explains the demonstrated policy better than alternative policies by a
margin.   The methods  under  this category  aim  to address  \irl{}'s
solution ambiguity (discussed in Section~\ref{subsection:accuracy}) by
converging  on a  solution  that maximizes  some  margin.  We  broadly
organize the methods  that engage in margin optimization  based on the
type of the margin that is used.

\subsubsection{Margin of optimal from other actions or policies} 

One of  the earliest and  simplest margins chosen for  optimization is
the  sum of  differences between  the  expected value  of the  optimal
action and that of the next-best action over all states,
\begin{equation}
\sum_{s\in S} Q^\pi(s,a^*)-\max_{a\in A\backslash \{a^*\}}Q^\pi(s,a)
\label{eqn:Q-margin}
\end{equation}
where $a^*$  is the optimal action for $s$.   

If  the reward  function  is  feature-based, whose  form  is given  in
Eq.~\ref{eqn:linear-reward},   a  similar   margin   that  takes   the
difference  between  the expected  value  of  the behavior  from  each
observed  trajectory  and  the  largest  of  the  expected  values  of
behaviors from all other trajectories can be used to learn the feature
weights. The expected value of a policy is obtained by multiplying the
empirical state visitation frequency from the observed trajectory with
the weighted  feature function values $\bm{\phi(\cdot)}$  obtained for
the trajectory. For each trajectory $\tau_i$ in the demonstration, the
margin can now be expressed as,
\begin{equation}
     \sum_{\langle s,a \rangle \in \tau_i}
     \psi(s)~\bm{w}^T\bm{\phi(s,a)} - \max_{\tau \in
       (S \times A)^l\backslash \{\tau_i\}}  \sum_{\langle s,a \rangle \in \tau}
     \psi(s)~\bm{w}^T\bm{\phi(s,a)}
\label{eqn:value-diff-MMP}
\end{equation}
where $(S \times A)^l$ is the set of all trajectories of length $l$.

An early and foundational method that  optimized the margin given in
Eq.~\ref{eqn:Q-margin} is Ng and
Russell's~\cite{Ng2000},  which  takes  in   the  expert's  policy  as
input. It formulates a linear  program to retrieve the reward function
that not  only produces the  given policy  as optimal output  from the
complete MDP, but  also maximizes the margin shown  above. In addition
to  maximizing this  margin,  it also  prefers  reward functions  with
smaller values as a form of regularization. 

Under the assumption that each  trajectory in a demonstration reflects
a distinct policy and the reward function is expressed as a linear and
weighted sum of feature functions, Ratliff et al.'s~\cite{Ratliff2006}
maximum  margin  planning  (\textsc{mmp}) associates  each  trajectory
$\tau_i \in \mathcal{D}$  with an MDP.  While these  MDPs could differ
in their  state and  action sets, and  the transition  functions, they
share  the same  reward function.   The desired  reward weight  vector
$\bm{w}$  is  obtained   by  solving  a  quadratic   program  that  is
constrained  to  have  a  positive   value  on  the  margin  given  in
Eq.~\ref{eqn:value-diff-MMP} with  the right-hand  side of  the margin
augmented by a regularizing loss term $l_i^T \psi$
that  quantifies  the closeness  between  the  demonstrated and  other
behaviors.  




\begin{figure}[ht!] 
	\centering{
	\includegraphics[scale=0.5]{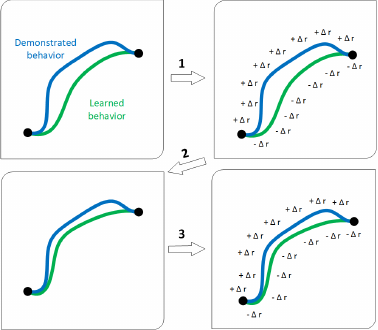}}
      \caption{An  iteration of \textsc{learch}  in the  feature space
        $\Phi=\{\phi(s,a)|\forall  (s,a)\in  S\times  A\}$ (Fig  3  in
        Ratliff et  al.~\cite{Ratliff2009} excerpted with permission).
        The method considers  a reward function as negative  of a cost
        function.  Blue path  depicts the demonstrated trajectory, and
        green  path  shows  the   maximum  return  (or  minimum  cost)
        trajectory  according  to   the  current  intermediate  reward
        hypothesis.  Step  1 is the determination of  the points where
        reward should be modified, shown as $-\Delta r$ for a decrease
        and  $+\Delta r$  for  an increase.   Step  2 generalizes  the
        modifications  to  entire  space  $\Phi$  computing  the  next
        hypothesis   $\hat{R}_E$  and  corresponding   maximum  return
        trajectory.      Next    iteration    repeats     these    two
        steps.}
\label{LEARCH_iteration}
\end{figure}
 
Using  the  same margin  as  in  Eq.~\ref{eqn:value-diff-MMP} and  the
regularizer, Ratliff  et al.~\cite{Ratliff2009} improves on  {\sc mmp}
in a subsequent method called {\em learn to search (\textsc{learch})}.
Figure~\ref{LEARCH_iteration}   explains    how   an    iteration   of
\textsc{learch}  increases the  cost  (decreases the  reward) for  the
actions  that cause  deviation  between the  learned and  demonstrated
behaviors.   For optimization,  \textsc{learch} uses  an exponentiated
functional  gradient   descent  in  the  space   of  reward  functions
(represented  as cost  maps). Later,  Silver et  al.~\cite{Silver2008}
introduced a  gradient normalization technique for  \textsc{learch} to
allow for suboptimal demonstrations as input.


\subsubsection{Margin of observed from learned feature expectations}

Adoption of the  feature-based reward function led  to several methods
optimizing margins that utilized  feature expectations.  Some of these
methods seek a  reward function that minimizes the  margin between the
feature   expectations   of  a   policy   computed   by  the   learner
(Eq.~\ref{eqn:feature-expectation})   and  the   empirically  computed
feature     expectations     from      the     expert's     trajectory
(Eq.~\ref{eqn:empirical-feature-expectation});
\begin{equation}
|\bm{{\mu}^{\phi}}(\pi)-\bm{\hat{\mu}^{\phi}}(\mathcal{D})|.
\label{eqn:feature-loss}
\end{equation} 
We refer to this margin as the {\em feature expectation loss}.

Two foundational  methods~\cite{Abbeel2004} that maximize  the feature
expectation    loss   margin    of   Eq.~\ref{eqn:feature-loss}    are
\textsc{max-margin} and \textsc{projection}.   Noting that the learner
does  not typically  have access  to the  expert's policy,  both these
methods take a demonstration (defined in Def.~\ref{def:irl}) as input.
The methods represent the reward function as a linear, weighted sum of
feature functions.

\begin{figure}[ht!] 
	\centering{
		\includegraphics[scale=1]{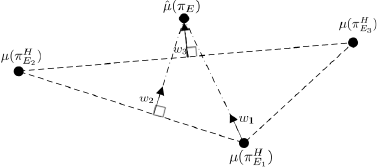}
		\caption{Iterations of the max-margin  method computing the weight
			vector,    $\mathbf{w}$,   and    the   feature    expectations,
			$\bm{\mu^{\phi}}$.    $\bm{\hat{\mu}^{\phi}}(\mathcal{D})$  is
			the       estimation      of       the      feature       counts
			$\bm{\mu^{ \mathbf{\phi}}}(\pi_E) $ of  the expert.  $ w_j $
			is  the learned  weight  vector in  the  $j^{th}$ iteration  and
			$\pi^H_{E_j}$   is   the   corresponding  optimal   policy   for
			intermediate  hypothesis. This  figure  is  redrawn with  slight
			changes from the one in Abbeel and Ng~\cite{Abbeel2004}.}
		\label{fig:maxmargin_iteration}
	}
\end{figure}

Both methods  iteratively tune weight  vector $\bm{w}$ by  computing a
policy as  an intermediate  hypothesis at  each step  and using  it to
obtain intermediate  feature counts.   These counts are  compared with
the empirical  feature counts  $\bm{\hat{\mu}^{\phi}}(\mathcal{D})$ of
expert    and    the    weights    are   updated,    as    shown    in
Fig.~\ref{fig:maxmargin_iteration}.   Abbeel and  Ng~\cite{Abbeel2004}
point  out that  the performance  of  these methods  is contingent  on
matching the  feature expectations,  which may  not yield  an accurate
$ \hat{R}_E $ because feature expectations are based on the policy. An
advantage of these methods is  that their sample complexity depends on
the number of features and not on the complexity of expert's policy or
the size of the state
space. 

A variant  of the projection method  described above is Syed  et al's.
multiplicative      weights      for      apprenticeship      learning
(\textsc{mwal})~\cite{Syed2008}. The  initial model and input  are the
same in both methods.  However,  \textsc{mwal} presents a learner as a
max  player choosing  a policy  and  its environment  as an  adversary
selecting  a  reward  hypothesis.   This  formulation  transforms  the
value-loss margin to  a minimax objective for a  zero-sum game between
the           learner          and           its          environment,
$                   \max_{\hat{\pi}_E}                   \min_{\bm{w}}
(\bm{w}^T\bm{\mu^{\phi}}(\hat{\pi}_E)-\bm{w}^T\bm{\hat{\mu}^{\phi}}(\mathcal{D}))
$,  and the  optimization uses  the exponentiated  gradient ascent  to
obtain the weights $\bm{w}$.

\subsubsection{Observed and learned policy distributions over actions} 
An alternative to  minimizing the feature expectation loss 
is to minimize the  probability difference between stochastic policies
\begin{equation}
\hat{\pi}_E (a|s) -  \pi_E(a|s)
\label{eqn:diff-policy-distribution}
\end{equation}
for each state. As the behavior  of expert is available instead of its
policy,  the  difference  above  is  computed  using  the  empirically
estimated state visitation  frequencies (Eq.~\ref{eqn:visit_freq}) and
the   frequencies  of   taking   specific  actions   in  the   states.
\textsc{hybrid}-\irl{}~\cite{Neu2007}                             uses
Eq.\ref{eqn:diff-policy-distribution}   in  the   margin  optimization
problem, solving the optimization using  gradient descent in the space
of reward hypotheses.



\subsection{Entropy Optimization}
\label{subsec:entropy-opt}

\irl{}  is essentially  an ill-posed  problem because  multiple reward
functions  can  explain the  expert's  behavior.   The maximum  margin
approaches  of Section~\ref{subsec:max-margin}  introduce a  bias into
the learned  reward function.   To avoid  this bias,  multiple methods
take recourse to the maximum entropy principle~\cite{Jaynes_MaxEnt} to
obtain  a  distribution over  behaviors,  parameterized  by the  reward
function weights.  According to  this principle, the distribution that
maximizes the entropy makes minimal commitments beyond the constraints
and is least  wrong.  We broadly categorize the  methods that optimize
entropy based on  the distribution, whose entropy is  being used, that
is chosen by the method.

\subsubsection{Entropy of the distribution over trajectories or policies}  

We may learn  a reward function that yields the  distribution over all
trajectories with the maximum entropy
\begin{equation}
\max_{\Delta}~ -\sum\limits_{\tau \in
		(S \times A)^l}Pr(\tau)\log Pr(\tau) 
\label{eqn:max-ent-traj}
\end{equation}
while being  constrained by the observed  demonstration.  However, the
search space  of trajectories  in this  optimization $(S  \times A)^l$
grows exponentially with  the length of the trajectory  $l$.  To avoid
this  disproportionate growth,  we may  learn a  reward function  that
alternately yields the distribution over all policies with the maximum
entropy
\begin{equation}
\max_{\Delta}~ -\sum\limits_{\pi \in
		(S \times A)}Pr(\pi)\log Pr(\pi) 
\label{eqn:max-ent-policy}
\end{equation}
where  $\Delta$ is  the space  of all  distributions. Notice  that the
space of policies grows with the sizes of the state and action sets as
$\mathcal{O}(|A|^{|S|})$ but not with the length of the trajectory.

A   foundational  and   popular   \irl{}  technique   by  Ziebart   et
al.~\cite{Ziebart2008}   \textsc{maxentirl}   optimizes  the   entropy
formulation   of    Eq.~\ref{eqn:max-ent-traj}   while    adding   two
constraints.  First, the distribution  over all trajectories should be
a probability distribution. Second, the  expected feature count of the
demonstrated                                              trajectories
$\sum_{\tau            \in            \mathcal{D}}Pr(\tau)\sum_{t=1}^l
\gamma^t\phi_k(s_t,a_t)$  must  match   the  empirical  feature  count
obtained using Eq.~\ref{eqn:empirical-feature-expectation}.


Mathematically, this  problem is  a convex but  nonlinear optimization
whose  Lagrangian  dual  reveals  that the  distributions  of  maximum
entropy  belong  to the  exponential  family.  Therefore, the  problem
reduces to finding the reward weights $\bm{w}$, which parameterize the
exponential distribution  over trajectories, which exhibits  the highest
likelihood of the demonstration, 
\begin{equation}
  \arg\max_{\bm{w}} 
  \sum_{\tau \in \mathcal{D}} log~Pr(\tau;
  \bm{w}) ~\text{where}~~ Pr(\tau;
    \bm{w}) \propto e^{\sum\limits_{\langle s,a \rangle \in \tau}
      \bm{w}^T \bm{\phi(s,a)}}.
\label{eqn:maxent-likelihood}
\end{equation}
We show this  distribution here as it is the  subject of other methods
as well. Ziebart et al. solves the Lagrangian dual, thereby maximizing
the likelihood, using gradient descent. 

Wulfmeier et al.~\cite{Wulfmeier2015} shows that the linearly-weighted
reward function in  \textsc{maxentirl} can be easily  generalized to a
nonlinear  reward  function  represented  by a  neural  network.   The
corresponding \textsc{deep maxentirl}  technique continues to maximize
the likelihood  of Eq.~\ref{eqn:maxent-likelihood} by using  its known
gradient  in  the  backpropagation  to  update  the  neural  network's
weights.

The optimization of  the likelihood of Eq.~\ref{eqn:maxent-likelihood}
is     also      the     subject      of     the      {\sc     pi-irl}
method~\cite{Aghasadeghi_continousspace},   which   generalizes   {\sc
  maxentirl} to continuous  state spaces. To enable  this, it replaces
the traditional feature  functions in a reward with  its path integral
formulations~\cite{Theodorou_pathintegral} that involves  not only the
known features, but  also the rate of change in  the continuous state,
and a  matrix giving the state  and goal costs.  {\sc  pi-irl} uses an
iteratively  sampling approach  to continually  improve the  region of
trajectory  sampling   and  the   path  integral  functions   used  in
determining the demonstrated trajectory rewards.

\begin{figure}[ht!] 
	\centering{
		\includegraphics[scale=1]{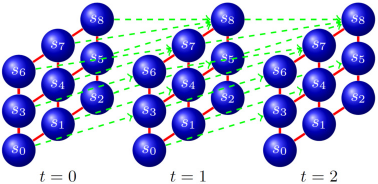}}
              \caption{A Markov random field
                which 
                favors  the policies which  prescribe  similar actions  in
                neighboring  states.  This  structure generalizes  the
                feature information beyond individual states.  (Figure
                reprinted  from  \cite{Boularias2012} with  permission
                from publisher.)
                \label{fig:Structured_MDP}}
\end{figure} 

Subsequent  to  \textsc{maxentirl},   Boularias  et  al.'s  structured
apprenticeship  learning~\cite{Boularias2012}  maximizes  the  entropy
shown  in  Eq.~\ref{eqn:max-ent-policy}  under  constraints  that  are
analogous to  those of  \textsc{maxentirl}. In particular,  the second
constraint matches  the expected  feature counts  of all  the policies
with     the     empirical     feature    count     obtained     using
Eq.~\ref{eqn:empirical-feature-expectation}.   The  resulting  convex,
nonlinear optimization problem is made somewhat tractable by observing
that   domains  often   exhibit   \textit{structure}  that   spatially
neighboring states have similar optimal  actions, which can be used to
guide  the  search.   This  results  in a  Markov  random  field  like
distribution      over      policies       as      illustrated      in
Fig.~\ref{fig:Structured_MDP}.

\subsubsection{Relative entropy of the distribution over trajectories} 

A  different  approach to  entropy  optimization  for \irl{}  involves
minimizing  the  relative  entropy  (also  known  as  Kullbach-Leibler
divergence~\cite{Kullback68informationtheory})       between       two
distributions $P$ and $Q$ over the trajectories. More formally,
\begin{equation} 
\min_{P\in \Delta} \sum_{\tau \in (S \times A)^l} P(\tau) \log \frac{P(\tau)}{Q(\tau)}.
\label{eqn:rel-entropy-objective}
\end{equation}
 
\textsc{reirl}~\cite{Boularias2011relative}  is a  prominent technique
that     utilizes    the     optimization    objective     given    in
Eq.~\ref{eqn:rel-entropy-objective}. Distribution  $Q$ in  {\sc reirl}
is obtained empirically by sampling trajectories under a {\em baseline
  policy}. Distribution $P$ is obtained such that the expected feature
count of the trajectories matches the empirical feature count obtained
using Eq.~\ref{eqn:empirical-feature-expectation}.  This constraint is
similar to  previous approaches  in this  section and  constrains {\sc
  reirl} to the  demonstration data.  The baseline policy  serves as a
way  to provide  domain-specific  guidance to  the  method.  While  an
analytical  solution   would  need  the  transition   dynamics  to  be
pre-specified,  Boularias  et al.   shows  that  the presence  of  the
baseline policy allows importance sampling  to be used and {\sc reirl}
can be solved model-free using stochastic gradient descent.

%




\subsection{Bayesian Update} 
\label{subsec:bayesian}

An important class of \irl{}  methods treats the state-action pairs in
a trajectory  as observations that  facilitate a Bayesian update  of a
prior  distribution over  candidate  reward  functions. This  approach
yields  a different  but principled  way for  \irl{} that  has spawned
various  methods. No  structure  is typically  imposed  on the  reward
function. Let $Pr(\hat{R}_E)$ be a  prior distribution over the reward
functions and  $Pr(\tau|\hat{R}_E)$ be  the observation  likelihood of
the reward  hypothesis. Then, a posterior  distribution over candidate
reward functions is obtained using the following Bayesian update:
\begin{equation}
         Pr(\hat{R}_E|\tau) \propto Pr(\tau|\hat{R}_E)~Pr(\hat{R}_E)
\label{eqn:bayes_update}
\end{equation}
Here,     the     likelihood      is     typically     factored     as
$Pr(\tau|\hat{R}_E)  = \prod\nolimits_{\langle  s,a \rangle  \in \tau}
Pr(\langle s,a \rangle |\hat{R}_E)$.  The update above is performed as
many times as the number of trajectories in the observed demonstration.

We categorize the Bayesian \irl{} methods  based on how they model the
observation likelihood in Eq.~\ref{eqn:bayes_update}.

\subsubsection{Boltzmann distribution} 

A popular choice for the likelihood function is the Boltzmann
distribution (also known as the Gibbs distribution) with the
Q-value of the state-action pair as the energy function:
\begin{equation}
P(\langle s,a \rangle|\hat{R}_E) \propto e^{\left(\frac{Q^*(s,a;\hat{R}_E)}{\beta}\right)}
\label{eqn:Boltzmann_model}
\end{equation} 
where parameter  $\beta$ controls  the randomness of the state-action
probability (lower the $\beta$, more random is the state-action pair).
Given a candidate reward hypothesis,  some state-action pairs are more
likely than others as given by the likelihood function.

The       earliest       Bayesian      \irl{}       technique       is
\textsc{birl}~\cite{Ramachandran2007}, which models  the likelihood as
shown in  Eq.~\ref{eqn:Boltzmann_model}.  {\sc birl}  suggests several
different priors over the continuous space of reward functions.  While
the uniform density function is an agnostic choice, several real-world
MDPs have a sparse reward structure  for which a Gaussian or Laplacian
prior is  better suited. If the  MDP reflects a planning  type problem
with  a large  dichotomy in  rewards (goal  states exhibiting  a large
positive  reward), then  a  Beta density  could  be appropriate.   The
continuous  space  of  reward  functions  brings  the  challenge  that
analytically  obtaining the  posterior is  difficult. To  address this
issue, {\sc birl}  presents a random walk based MCMC  algorithm in the
space of policies for obtaining a sample-based empirical approximation
of  the posterior.  On the  other hand,  we may  also converge  to the
maximum-a-posteriori   reward   function   directly   using   gradient
descent~\cite{Choi11:MAP}.

Lopes  et al.~\cite{Lopes2009}  show  that the  posterior computed  in
\textsc{birl} can  be used to  incorporate active learning  in \irl{}.
The learner  queries the expert  for additional samples of  actions in
those states where a distribution  induced from the posterior exhibits
a  high  entropy.   The  induced  distribution is  a  measure  of  how
discriminating is  the learned  expert policy at  each state.  The new
samples help learn an improved posterior.

\subsubsection{Gaussian process}

An influential Bayesian approach introduced  after {\sc BIRL} lets the
reward    function   be    a   nonlinear    function   of    features,
$\hat{R}_E=f(\bm{r},\bm{\phi})$, and  models it as a  Gaussian process
whose underlying structure is given by a kernel function parameterized
by  $\bm{\theta}$.    Thus,  the  posterior   $Pr(\hat{R}_E|\tau)$  in
Eq.~\ref{eqn:bayes_update}                 now                 becomes
$Pr(\bm{r},  \bm{\theta} |  \tau, \bm{\phi})$  where $\bm{r}$  are the
rewards associated  with the  feature functions $\bm{\phi}$  at select
states and actions.

The {\sc gpirl} technique~\cite{Levine2011} computes this posterior as
a      Bayesian     update      with     the      likelihood     being
$Pr(\tau|\hat{R}_E)Pr(\hat{R}_E|\bm{r},\bm{\theta},  \bm{\phi})$.  The
first    factor     is    computed    as    shown     previously    in
Eq.~\ref{eqn:Boltzmann_model}. However, the  distinctive second factor
(also   called  the   Gaussian  process   posterior)  is   a  Gaussian
distribution   with   analytically   derived   mean   and   covariance
matrices. {\sc GPIRL}  generalizes from just a small  subset of reward
values  -- those  contained in  the  observed trajectories  and a  few
additional rewards at random states.  It utilized L-BFGS with restarts
to  optimize the  likelihood,  finding  most likely  $  \bm{r} $  and,
thereby, most likely reward functions.

\subsubsection{Maximum likelihood estimation} 

Apart from the  posterior estimation in \textsc{birl}  methods, we may
just   directly   maximize   the   likelihood  of   the   input   data
($Pr(\tau|\hat{R}_E)$  in  Eq.~\ref{eqn:bayes_update}).  The  standard
expression for the data likelihood involves the policy and the Bellman
operator, which  is not  differentiable. However, a  softmax Boltzmann
exploration  can  be  used  to   change  the  policy  expression  from
maximization                                                        to
$  \pi_{\bm{w}}(s,a)=\frac{e^{\beta   Q(s,a;  \hat{R}_E)}}{\sum_{a'\in
    A/a}  e^{\beta  Q(s,a';\hat{R}_E)}}$,  which  in  turn  makes  the
likelihood    $P(\tau|\hat{R}_E)$     differentiable.     Vroman    et
al.~\cite{Babes-Vroman2011} chooses this  Boltzmann exploration policy
to  infer the  reward feature  weights $  \bm{w} $  that leads  to the
maximum likelihood estimate.  As the likelihood is now differentiable,
the algorithm  uses a standard  gradient ascent for converging  to the
(locally-) optimal weights.

\subsection{Classification and Regression}
 
Classical  machine  learning  techniques such  as  classification  and
regression have  also played a  significant role in  \irl{}.  However,
these  methods  are challenged  by  the  fact  that  \irl{} is  not  a
straightforward supervised learning problem. Nevertheless, the methods
below show that \irl{} can  be cast into this framework. 


\subsubsection{Classification based on action-value scores} 

\irl{} may  be formulated as  a multi-class classification  problem by
viewing  the   state-action  pairs  in  a   trajectory  as  data-label
pairs. For each state in a pair, the label is the corresponding action
performed at that state as prescribed by the expert's policy. As there
are usually more than two  actions in most domains, the classification
is  into  one  of  multiple  classes. An  obvious  way  to  score  the
classification is to  use the action-value function,  which is derived
from Eq.~\ref{eqn:value-function-feature} as:
\begin{equation}
Q^{\pi}(s,a) = \bm{w}^T~\bm{\mu}^{\phi}(\pi)(s,a).
\label{eqn:score-scirl}
\end{equation}
Notice that this a linear scoring function which uses the same feature
weight vector $\bm{w}$ as used in the reward function. Subsequently, a
classifier aims to learn the  weights that minimize the classification
error between the labels in the state-action pairs of a trajectory and
the action label predicted as $\arg\max_{a \in A} Q^\pi(s,a)$.


Klein  et al.~\cite{Klein2012structured}  introduced this  multi-class
classification   formulation   of   \irl{}    in   a   method   called
\textsc{scirl}.   The   demonstration  is  utilized  for   training  a
classifier  with Eq.~\ref{eqn:score-scirl}  as  the scoring  function.
Any  linear score  based multi-class  classification algorithm  may be
used to solve  for the weight vector $\bm{w}$.  {\sc  scirl} chose the
large          margin          approach         of          structured
prediction~\cite{taskar2005learning}    as     the    algorithm    for
classification.

While  {\sc scirl}  assumed  the  presence of  a  transition model  to
compute     the    scoring     function,    an     extension    called
\textsc{csi}~\cite{Klein_CSI} takes  a step further and  estimates the
transition probabilities if they are not known.  \textsc{csi} utilizes
standard regression on a simulated  demonstration data set to estimate
the transition  model and thereafter  learn the reward  function using
{\sc scirl}.

Notice that  {\sc scirl} assumes  that the action in  the state-action
pair is the desired optimal action  and uses it as the label. However,
if   we  allow   for  the   demonstration  to   be  suboptimal,   this
classification  approach may  not  work.  In  this  context, Brown  et
al.~\cite{brown19a} show that  if the input to \irl{}  also includes a
ranking of  the trajectories based  on the degree of  suboptimality of
the trajectory, we may  utilize this additional preference information
to train  a neural  network representation of  the reward  function. A
cross-entropy based loss function trains  the neural network to obtain
a higher cumulative  reward for the trajectory that  is preferred over
another per the given ranking.

\subsubsection{Regression tree for state space partitions} 

The linearly  weighted sum  of feature  functions represents  a global
reward function model over the entire  state and action spaces. It may
be inadequate or require many features when these spaces are large. An
alternate model  could be a  regression tree whose  intermediate nodes
are  the individual  feature functions  and whose  leaves represent  a
conjunction of  indicator feature functions.   Each path of  this tree
then captures  a region of the  state and action space,  and the whole
tree induces a partition of this space.
  
Subscribing   to  this   representation   of   the  reward   function,
\textsc{firl}~\cite{Levine2010_featureconstruction}        iteratively
constructs both the features and the regression tree. To arrive at the
reward values for the current regression tree, it solves a quadratic
program with the objective function 
$$ \min_{\hat{R}_E} 
|| \hat{R}_E  - Proj_{\bm{\phi}^{(i-1)}} (\hat{R}_E) ||_2$$  under the
constraint that  the policy obtained  by solving the MDP  with current
$\hat{R}_E$ gives actions  that match those in  the demonstrations for
the associated states.  Here, $Proj_{\bm{\phi}^{(i-1)}}(\cdot)$ is the
projection of the reward function on the linear combination of the set
of  features $\bm{\phi}^{(i-1)}$  from  the  previous iteration.  {\sc
  firl} interleaves this optimization step  with a fitting step during
which a new set of  features $\bm{\phi}^{(i)}$ is learned by splitting
those leaves  of the tree are  that too coarse or  merging leaves that
yield the  same average reward  values.  The iterations stop  when the
learner detects that further node splitting is unnecessary to maintain
consistency with the demonstration.

\begin{table}[ht!]
	{\scriptsize
		\begin{center}
			\bgroup
			\def\arraystretch{1.2}
			\begin{tabular}{ | l | l |}
				\hline 
				$\bm{\phi}(\cdot)$ & 
				{reward features}  \\ \hline
                               	$ \bm{w} $ & 
				{reward feature weights}  \\ \hline
				$ \psi^{\pi}(s) $ & 
				{state visitation frequency}  \\ \hline
				$\bm{\mu}^{\phi}(\pi)$ & 
				{feature expectations}  \\ \hline
				$ \tau $ & 	{trajectory}  \\ \hline
				$ \mathcal{D} $ & 	{demonstration
                                                    (set of trajectories)}  \\ \hline
				$ \beta $ & 	{temperature in Boltzmann distribution}  \\ \hline
				$ M $ & {number of target reward functions} \\ \hline
				{IBP} & {indian buffet Gaussian
                                        process} \\ \hline
			\end{tabular}
			\caption{A   compilation   of   the   notation
                          introduced  in the  text and  abbreviations,
                          which        is         referenced        in
                          Tables~\ref{tbl:challengesAddressedByFoundMethods}
                          and~\ref{tbl:challengesAddressedByExtensions}.}
			\label{tbl:notations1}
			\egroup
		\end{center}}
	\end{table}

\subsection{Summary of the Methods}

In  the  previous subsections,  we  briefly  described the  early  and
foundational \irl{} methods.  These have influenced various subsequent
methods and spawned  improvements as discussed in  the later sections.
Table~\ref{tbl:challengesAddressedByFoundMethods}     abstracts    and
summarizes  the  key insights  of  these  methods. It  identifies  the
parameters  that are  learned,  the metric  used  in the  optimization
objective,  and  a distinguishing  contribution  of  the method.  This
facilitates a convenient comparison across the techniques and helps in
aligning    the     methods    with    the    template     given    in
Algorithm~\ref{alg:IRL_template}.

	\begin{table}[ht!]
		\begin{center}
			{\scriptsize
				\bgroup
				\def\arraystretch{1.2}
				\begin{tabular}{ | l || c | c| l |}
					\hline
					\textbf{Method} &
					\textbf{$\hat{R}_E$ params} &
					\textbf{Optimization objective} & 
					\textbf{Notable aspect} \\ \hline	
                                  \multicolumn{4}{|l|}{{\bf Max
                                  margin methods} - maximize the margin between value of observed
                                  behavior and the hypothesis}
                                  \\ \hline                               
                                  \textsc{mmp} 
                                                        &  \multirow{6}{*}{$\bm{w}$} &
                                                                                       value
                                                                                       of
                                                                                       obs. $\tau$
                                                                                       -
                                                                                       max
                                                                                       of
                                                                                       values
                                                                                       from
                                                                                       all
                                                                                       other
                                                                                       $\tau$
                                                                                       ~(Eq.~\ref{eqn:value-diff-MMP})
                                                                
%
                                  &
                                  provable convergence   \\  \cline{1-1}\cline{3-4} 
                                  \textsc{max-margin} &  & feature
                                                           exp. of
                                                           policy -
                                                           empirical
                                                           feature
                                                           exp. (Eq.~\ref{eqn:feature-loss})
                                  &
                                  sample bounds   \\  
                                  \cline{1-1}\cline{3-4} 
                                                                    \textsc{mwal} & & min
                                                  diff. in value
                                                  of policy and
                                                  observed $\tau$
                                                  across features  
                                  &
                                  first bound on \\   
                                   &  & & iteration complexity 
                                  \\ \cline{1-1}\cline{3-4}

                                  \textsc{hybrid-irl} &  &  empirical stochastic
                                                           policy -
                                                           computed
                                                           policy of
                                                           expert (Eq.~\ref{eqn:diff-policy-distribution}) 
                                  
                                  & natural
                                  gradients and \\ 
                                  & & & efficient optimization \\
                                  \cline{1-1}\cline{3-4} \hline
                                  \textsc{learch} &  $R(\bm{\phi})$ & \multirow{4}{*}{value
                                                                                       of
                                                                                       obs. $\tau$
                                                                                       -
                                                                                       max
                                                                                       of
                                                                                       values
                                                                                       from
                                                                                       all
                                                                                       other
                                                                                       $\tau$
                                                                                       ~(Eq.~\ref{eqn:value-diff-MMP})}
                                  &
                                  nonlinear
                                  reward with\\ 
                                  & & & 
                                  suboptimal
                                  input
                                  \\ \cline{1-1}\cline{4-4}
				{Silver et al.~\cite{Silver2008}} &  &  &  normalization of  \\ 
					 & & & outlier inputs \\  \hline
                                  \multicolumn{4}{|l|}{{\bf Max
                                  entropy methods} - maximize the entropy of the distribution over behaviors} \\ \hline 
                                  \textsc{maxentirl} &
                                                       \multirow{6}{*}{$\bm{w}$}
                                                                    &  
                                  entropy of distribution over
                                                                      trajectories (Eq.~\ref{eqn:max-ent-traj})                                    
                                  & low learning bias \\
                                  \cline{1-1}\cline{3-4}
                                  \textsc{structured} &
                                                                    &
                                                                      entropy
                                                                      of
                                                                      distribution
                                                                      over
                                                                      policies
                                                                      (Eq.~\ref{eqn:max-ent-policy}) 
                                  & efficient optimization \\ 
                                  \textsc{apprenticeship} & & & \\ \cline{1-1}\cline{3-4}

                                  	\textsc{deep maxentirl} &   &
                                                                       \multirow{3}{*}{gradient
                                                                       of
                                                                       likelihood
                                                                       equivalent
                                                                       of
                                                                      MaxEnt (Eq.~\ref{eqn:maxent-likelihood})}
					& nonlinear reward \\
                                  \cline{1-1}\cline{4-4}
      					\textsc{pi-irl}  & & &  continuous \\ 
					&   &  &  state-action spaces\\ \cline{1-1}\cline{3-4}
                                  
                                  \textsc{reirl} &   & relative
                                                       entropy of
                                                       distribution
                                                       from baseline
                                                       policy (Eq.~\ref{eqn:rel-entropy-objective})
                                  &  suboptimal input and \\ 
                                  & & & unknown dynamics \\ \hline

					\multicolumn{4}{|l|}{{\bf Bayesian
                                  learning methods} - learn posterior over hypothesis space using Bayes rule} \\ \hline 
					\textsc{birl} &
                                                        \multirow{3}{*}{$R(s)$}
                                                                    &
                                                                      posterior
                                                                      with
                                                                      Boltzmann
                                                                      data
                                                                      likelihood (Eq.~\ref{eqn:Boltzmann_model})
					& first Bayesian  \\
					& & & \irl{} formulation \\
                                  \cline{1-1}\cline{3-4}
					{Lopes et
                                  al. \cite{Lopes2009}} &  & entropy
                                                             of
                                                             multinomial$(p_1(s),p_2(s),\ldots,p_{|A|-1}(s))$ 
					& active learning \\ 
                                        & &  derived from posterior &
                                  \\ \hline
                          		\textsc{gp-irl} & $f(\bm{r,\theta})$
                                        & Gaussian process posterior
                                        &  nonlinear reward \\ \hline
                                        \textsc{mlirl} &  $\bm{w}$
                                                                    &
                                                                      differentiable
                                                                      likelihood
                                                                      with
                                                                      Boltzmann
                                                                      policy (Eq.~\ref{eqn:Boltzmann_model})
                                                                   &
                                                                     first
                                                                     ML
                                  approach \\ \hline
		
					\multicolumn{4}{|l|}{{\bf Classification and regression} - learn a prediction model that imitates observed behavior} \\ \hline 
					\textsc{scirl} &
                                                         \multirow{3}{*}{$\bm{w}$}
                                                                    &
                                                                      \multirow{3}{*}{Q-function
                                                                      as
                                                                      classifier
                                                                      scoring
                                                                      function}
                                                                   &
                                                                     actions
                                                                     as state labels \\ 
					& & & provable
                                              convergence \\ \cline{1-1}\cline{4-4}
					\textsc{csi} & & & unknown dynamics \\ \hline 
					\textsc{firl} & regression &
                                                                     norm
                                                                     of
                                                                     ($\hat{R}_E$
                                                                     -
                                                                     projection
                                                                     of $\hat{R}_E$)
                                                                   &  avoids manual \\ 
					& tree &  & feature engineering\\ 
					\hline 
				\end{tabular}
				\caption{A  categorized  summarization
                                  of    the    foundational    methods
                                  presented                         in
                                  Section~\ref{sec:PrimaryMethods}.
                                  We focus on the  key aspects of each
                                  method and  abstract out  the shared
                                  representations.     For    example,
                                  notice    how    popular   is    the
                                  linearly-weighted  representation of
                                  the   reward  function.    Refer  to
                                  Table~\ref{tbl:notations1}   for   a
                                  quick  explanation of  abbreviations
                                  and notations used here.}
				\label{tbl:challengesAddressedByFoundMethods}
			}
			\egroup
		\end{center}
	\end{table}

\section{Mitigating the Challenges}
\label{sec:Mitigation} 

Next,  we   elaborate  how   the  foundational  methods   reviewed  in
Section~\ref{sec:PrimaryMethods} mitigate  the various  challenges for
\irl{}    introduced   in    Section~\ref{sec:challenges}.   Technical
challenges often drive the development of methods.  Hence, in addition
to situating the overall progress of the field, this section will help
the reader make  an informed choice about the method  that may address
the challenges in her particular domain.
We have also included techniques  that purposefully extend some of the
foundational methods to address a specific challenge.

\subsection{Improving the Accuracy of Inference}
\label{subsection:achieving_accuracy} 

As   we  mentioned   in  Section~\ref{subsection:accuracy},   \irl{}'s
inference  accuracy  depends on  several  components  of the  learning
process.   Most existing  methods aim  at ensuring  that the  input is
accurate, reducing  the ambiguity among multiple  solutions, improving
feature selection, and offering algorithmic performance guarantees.

\subsubsection{Learning from Noisy Input} 

Perturbed demonstrations  may be  due to  noisy sensing  or suboptimal
actions by the expert.
Methods such as {\sc  reirl} stay robust to perturbations whereas
other     \irl{}    methods     may    learn     inaccurate    feature
weights~\cite{Abbeel2007}       or       predict      the       action
poorly~\cite{Ratliff2006}. 
Methods such as \textsc{maxentirl}, \textsc{birl}, \textsc{mlirl}, and
\textsc{gpirl}  use  probabilistic  frameworks   to  account  for  the
perturbation.  For example, \textsc{mlirl}  allows tuning of its model
parameter  $\beta$  in  Eq.~\ref{eqn:Boltzmann_model}  to  allow  more
randomness into the learned policy $\hat{\pi}_E$ when the demonstrated
behavior       is      expected       to       be      noisy       and
suboptimal~\cite{Babes-Vroman2011}.  On  the other hand,  methods such
as \textsc{mmp} and \textsc{learch} introduce slack variables in their
optimization  objective for  this purpose.   Using the  application of
helicopter flight control, Ziebart et al.~\cite{ZiebartBD_compare_MMP}
show  that  the  robustness  of  \textsc{maxentirl}  to  an  imperfect
demonstration is better than that of \textsc{mmp}.
The    recent    method    by    Brown   et    al.     (called    {\sc
  t-rex})~\cite{brown2018efficient}  was  shown   to  learn  a  reward
function from sub-optimal input, which then surpasses the performances
of the demonstrator in simulated domains.

To specifically  address noisy input, Coates  et al.~\cite{Coates2008}
introduce  a   model-based  technique  of  trajectory   learning  that
de-noises the noisy demonstration  by learning a generative trajectory
model    and    then    utilizing    it    to    produce    noise-free
trajectories. Apprenticeship  learning is subsequently applied  to the
resultant  noise-free  but unobserved  trajectories~\cite{Coates2009}.
Melo   et   al.~\cite{Melo_perturbedinput}   formally   analyzed   and
characterized the space of solutions for the case when some actions in
a demonstration  are not optimal  and when the demonstration  does not
include samples  in all states.  Such demonstrations were  obtained by
perturbing the distribution that modeled the expert's policy.
Taking a step  further,  Shiarlis et  al.~\cite{Shiarlis_2016_failed}
performs \irl{} with some demonstrations that fail to even complete the
task.


\begin{figure}[ht!] 
	\centering{
		\includegraphics[scale=1.4]{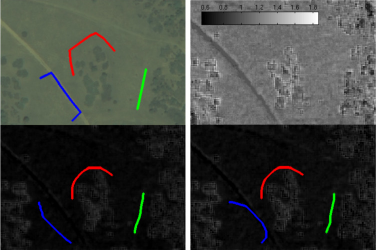}
                \caption{Learning with a perturbed demonstration in an
                  unstructured             terrain.             Figure
                  reprinted~\cite{Silver2008} with permission from MIT
                  press.   An  expert   provides  three  demonstration
                  trajectories - red, blue, and green [top left].  The
                  portion   of  terrain   traveled  by   a  presumably
                  achievable red trajectory should have low cost (high
                  reward) as  the expert  is presumably  optimal.  But
                  the path is not optimal.   It is not even achievable
                  by  any  planning  system with  predefined  features
                  because passing through the  grass is always cheaper
                  than  taking a  wide berth  around it.   The assumed
                  optimality   of  expert   forces  the   optimization
                  procedure  in  \irl{}  methods  to  lower  the  cost
                  (increase the reward) for features encountered along
                  the path, i.e., features for grass.  This influences
                  the  learning behavior  in other  paths such  as the
                  blue  path   [bottom  left].   Using   a  normalized
                  functional gradient~\cite{Silver2008}  mitigates the
                  lowering of costs [bottom right].}
                \label{learning_faultyinput}
		}
\end{figure} 
		
Suboptimal demonstrations may also  include trajectories whose lengths
are    much   longer    than   expected.     As   we    mentioned   in
Section~\ref{subsec:max-margin},  \textsc{mmp} minimizes  the cost  of
simulated trajectories diverging from  the demonstrated ones by noting
the difference in state-visitation frequencies of two trajectories.
\textsc{mmp} attempts this minimization for a suboptimal demonstration
as  well,  but avoids  it  if  the  learning method  distinguishes  an
unusually long demonstration from the optimal ones.
Silver et  al.~\cite{Silver2008,Ratliff2009} specifically  target this
issue  by   implementing  an  \textsc{mmp}-based   imitation  learning
approach  that  applies  a   functional  gradient  normalized  by  the
state-visitation    frequencies   of    a   whole    trajectory   (see
Fig.~\ref{learning_faultyinput} for an illustration).

\subsubsection{Ambiguity and Degeneracy of Reward Hypotheses}

Several methods mitigate this challenge of ambiguity and degeneracy by
better  characterizing the  space of  solutions.  This  includes using
heuristics  and  prior  domain   knowledge,  and  adding  optimization
constraints.

\textsc{mmp}  and \textsc{mwal}  avoid degenerate  solutions by  using
heuristics that favor the learned  value $V^{\hat{\pi}_E}$ to be close
to expert's $V^{\pi_E}$.  Specifically, \textsc{mmp} avoids degeneracy
by using  a loss function, which the degenerate $\hat{R}_E=0$ can not minimize because
the     function      is     proportional      to     state-visitation
frequencies~\cite{Neu2008}.
\textsc{hybrid}-\irl{}   avoids  degeneracy   in  the   same  way   as
\textsc{mmp}, and  makes the solution  less ambiguous by  preferring a
reward function that corresponds  to a stochastic policy $\hat{\pi}_E$
with      action     selection      same      as     the      expert's
($\hat{\pi}_E(a|s)  \approx  \pi_E(a|s)$).   Naturally, if  no  single
non-degenerate  solution makes  the  demonstration optimal,  ambiguous
output   cannot    be   entirely    avoided   using    these   methods
\cite{Ziebart2010}.

		
On the other  hand, Bayesian and entropy  optimization methods embrace
the ambiguity by modeling the  uncertainty of the hypothesized rewards
as a probability  distribution over reward functions or  that over the
trajectories   corresponding  to   the  rewards.    In  this   regard,
\textsc{maxentirl}  infers  a  single   reward  function  by  using  a
probabilistic framework  that avoids any constraint  other than making
the value-loss zero, $V^{\hat{\pi}_E}=V^{\pi_E}$.
On  the other  hand,  the maximum-a-posteriori  objective of  Bayesian
inference techniques and \textsc{gpirl}  limit the probability mass of
the posterior distribution to the  specific subset of reward functions
that supports  the demonstrated behavior.  This  change in probability
mass  shapes the  mean  of the  posterior, which  is  output by  these
methods.
Active  learning of  the  reward function  uses the  state-conditional
entropy   of   the  posterior   to   select   the  least   informative
states~\cite{Lopes2009}  and query  for further  information in  those
states.  The  selection mechanism builds on  \textsc{birl} and reduces
the  ambiguity  of the  solution  as  compared to  \textsc{birl}.   In
general, these methods add optimization constraints and exploit domain
knowledge to distinguish between the multiple hypotheses.\\

\noindent  We  believe that  the  progress  made collectively  by  the
methods in significantly  mitigating this challenge of  \irl{} -- that
it  is an  underconstrained  learning problem  represents  a key  {\em
  milestone} in the progression of this relatively new field.

  The presence of degenerate and  multiple solutions led early methods
  such as  \textsc{max-margin} and \textsc{mwal} to  introduce bias in
  their optimizations.  However, a  side effect of  this bias  is that
  these  methods   may  compute  a  policy   $\hat{\pi}_E$  with  zero
  probability    assigned     to    some    of     the    demonstrated
  actions~\cite{Ziebart2010}. Indeed, this is also observed in maximum
  likelihood  based  approaches  such as  \textsc{mlirl}.   Subsequent
  methods have  largely solved this issue.   For example, \textsc{mmp}
  makes the  solution policy have state-action  visitations that align
  with   those  in   the   expert's  demonstration.    \textsc{maxent}
  distributes  probability  mass  based   on  entropy  but  under  the
  constraint of feature expectation matching.  Further, \textsc{gpirl}
  addresses it  by assigning a  higher probability mass to  the reward
  function  corresponding   to  the  demonstrated   behavior,  seeking
  posterior distributions with low variance.


\subsubsection{Theoretical Bounds on Accuracy}

From  a theoretical  viewpoint, some  methods have  better performance
guarantees than others.  The  maximum entropy probability distribution
over  space of  trajectories  (or policies)  minimizes the  worst-case
expected  loss \cite{Grunwald2004}.   Consequently, \textsc{maxentirl}
learns a behavior which is neither much better nor much worse than the
expert's  \cite{Dimitrakakis2012}.  However,  the worst-case  analysis
may not represent the performance  in practice because the performance
of optimization-based  learning methods can be  improved by exploiting
favorable properties of the  application domain.  Classification based
approaches such as \textsc{csi} and \textsc{scirl} admit a theoretical
bound for  the quality of  $\hat{R}_E$ in  terms of optimality  of the
learned  behavior $\hat{\pi}_E$,  given that  both classification  and
regression  errors are  small.   Nevertheless, these  methods may  not
reduce the  loss as much  as \textsc{mwal} as  the latter is  the only
method, in  our knowledge, which  has no  lower bound on  the incurred
value-loss~\cite{Syed2008}.

Some  methods also  analyze  and  bound the  ILE  metric  for a  given
threshold of  success and  a given  minimum number  of demonstrations.
The analysis  relies on determining  the value  of a policy  using its
generated                     feature                     expectations
$\mathbf{\mu}^{\mathbf{\phi}}(\pi)$~\cite{Abbeel2004,Ziebart2008,Choi2011}
or                     state-visitation                    frequencies
$\psi^{\pi}(s)$~\cite{Neu2007,Ziebart2008,Ratliff2006,Babes-Vroman2011}
as shown in Eq.~\ref{eqn:value-function-feature}.

For  any  method based  on  feature  expectations or  state-visitation
frequencies, there exists  a probabilistic upper bound on  the bias in
$\hat{V}^{\pi_E}$  and  thereby on  ILE  for  a given  minimum  sample
complexity~\cite{Abbeel2004,Syed2008_supplement,Vroman2014}.     These
bounds     apply     to      methods     such     as     \textsc{mmp},
\textsc{hybrid}-\textsc{irl},   and    \textsc{maxentirl}   that   use
state-visitation frequencies.  Subsequently, the derived bound on bias
can be used to analytically compute  the maximum error in learning for
a given minimum sample complexity
Lee  et al.~\cite{Lee_Improved_Projection}  change the  criterion (and
thereby  the   direction)  for   updating  the  current   solution  in
\textsc{max-margin} and \textsc{projection}  methods to formally prove
an improvement in the accuracy of  the solution as compared to that of
the  original  method.  A  recent  extension  of BIRL  introduced  the
bounding of approximated ILE as an  alternative to the bounding of the
difference     in    feature     expectations     as    a     learning
objective~\cite{brown2018efficient}.     The    method    demonstrated
confidence error bounds tighter than  the methods that used the latter
objective.

\subsection{Generalizability}
\label{subsection:generalization_improvement}

While   early  approaches   such  as   apprenticeship
  learning  required a  demonstration that  spanned all  states, later
  approaches  sought  to  explicitly  learn  a  reward  function  that
  correctly represented  expert's preferences for  unseen state-action
  pairs, or  one that is valid  in an environment that  mildly differs
  from the  input.  An added  benefit is  that such methods  may need
less demonstrations.
\textsc{gpirl} can learn the reward for unseen states lying within the
domains  of   the  features  of  a   Gaussian  process.   Furthermore,
\textsc{firl} can  use the learned  reward function in  an environment
that  is slightly  different from  the original  environment used  for
demonstration but with base features  similar to those in the original
environment.    Similarly,  Finn   et  al.'s   guided  cost   learning
(\textsc{gcl})~\cite{Finn_gcl}, which extends {\sc reirl} to include a
neural network representation  for the reward function  and a baseline
distribution learned  using RL,  admits an enhanced  generalization by
learning new  instances of a previously-learned  task without repeated
reward learning.  Melo and Lopes~\cite{S.Melo2010} shows  that the use
of  bisimulation  metrics  allows \textsc{birl}  to  achieve  improved
generalization  by partitioning  the state  space based  on a  relaxed
equivalence between the states.

\begin{figure}[ht!] 
  \centering{
    \includegraphics[scale=1]{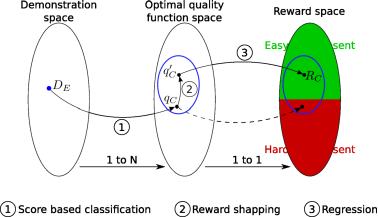}}
  \caption{ The  3 step  process of  \irl{} in  a relational  domain -
    classification outputs a score  function, reward shaping optimizes
    the  output,  and  regression   generates  an  approximate  reward
    function   $\hat{R}_E$   corresponding   to  the   optimal   score
    function. Included with permission from authors.}
  \label{fig:CSI}
\end{figure} 
	

Munzer    et    al.      \cite{Munzer_relationalIRL}    extends    the
classification-regression steps in  \textsc{csi} to include relational
learning  in  order to  benefit  from  the strong  generalization  and
transfer  properties  that  are  associated  with  relational-learning
representations.   The process  shapes the  reward function  using the
scoring   function   as   computed    by   the   classification   (see
Fig.~\ref{fig:CSI} for more details).

\subsection{Lowering Sensitivity to Prior Knowledge} 
In this section, we discuss techniques in the context of the challenge
introduced  in   Section  \ref{subsection:sensitivity_priorknowledge}.
Performance of  the foundational methods such  as \textsc{projection},
\textsc{max-margin}, \textsc{mmp}, \textsc{mwal}, \textsc{learch}, and
\textsc{mlirl} are all highly sensitive  to the selection of features.
While we are  unaware of methods that explicitly seek  to reduce their
dependence on  feature selection,  some methods  are less  impacted by
virtue of  their approach.  These include  \textsc{hybrid}-\irl{} that
uses  policy  matching and  all  maximum  entropy based  methods  tune
distributions  over the  trajectories or  policies, which  reduces the
impact   that   feature   selection   has  on   the   performance   of
\irl{}~\cite{Neu2008}.

Apart from  selecting appropriate  features, the  size of  the feature
space influences  the error  in learned  feature expectations  for the
methods  that rely  on  $\hat{V}^{\pi_E}$, e.g.,  \textsc{projection},
\textsc{mmp},  \textsc{mwal},  and  \textsc{maxentirl}.  If  a  reward
function is linear taking  the form of Eq.~\ref{eqn:linear-reward} and
the value of each of its $k$  features is bounded from the above, then
the   probable  bound   on  the   error  scales   linearly  with   $k$
\cite{Ziebart2008}.  However,  maximum entropy  based methods  show an
improvement in this aspect with $O(\log k)$ dependence.
	
\subsection{Analysis and Reduction of Complexity}
\label{subsection:lowering_learningcost} 


The intractability of  this machine learning problem due  to its large
hypothesis  space   has  been  significantly  mitigated   through  the
widespread adoption of a reward function composed of linearly-weighted
features. Though  this imposed structure limits  the hypothesis class,
it often  adequately represents  the reward  function in  many problem
domains. Importantly, it allowed the  use of feature expectations as a
sufficient statistic for representing the value of trajectories or the
value of  an expert's policy.   This has contributed  significantly to
the success of early methods such as \textsc{projection}, \textsc{mmp}
and \textsc{maxentirl}. Consequently, we view  this adoption as a {\em
  milestone} for this field.

Next, we  discuss ways by which  \irl{} methods have sought  to reduce
the time and space complexity of  an iteration, and mitigate the input
sample   complexity.    Early   \irl{}   methods  such   as   Ng   and
Russell~\cite{Ng2000} and {\sc projection} were mostly demonstrated on
grid  problems  exhibiting  less  than   a  hundred  states  and  four
actions. Later methods based on  entropy optimization and {\sc GP-IRL}
scaled   up  with   {\sc   maxentirl}  demonstrating   results  on   a
deterministic MDP  with thousands of  states and actions  while taking
recourse to approximations.
		
\begin{figure}[ht!] 
\centering{
\includegraphics[scale=1]{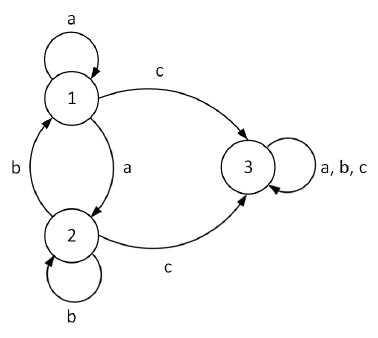}}
\caption{State equivalence in an  \mdp{} with states $S=\{1,2,3\}$ and
  actions $A=\{a,b,c\}$. The similarity in actions and transitions for
  states 1 and  2 makes them equivalent.  Therefore,  the selection of
  optimal actions through  expert's policy $\pi_E$ will  be similar in
  both the states. Demonstration of  $c$ in one implies the optimality
  of $c$  in other.  The illustration  redraws Fig. 1 in  Melo et
  al. \cite{S.Melo2010}.}
\label{fig:stateequivalence_bisimulation}
\end{figure}

While  an  emphasis on  reducing  the  time  and space  complexity  is
generally lacking among \irl{} techniques, a small subset does seek to
reduce the time complexity.
An analysis of \textsc{birl} shows that computing the policy $ \pi_E $
using the mean  of the posterior distribution  is computationally more
efficient than the direct minimization of expected value-loss over the
posterior~\cite{Ramachandran2007}.   Specifically,  the  Markov  chain
with  a  uniform  prior   that  approximates  the  Bayesian  posterior
converges   in   polynomial   time.    Enhancing   \textsc{birl}   via
bisimulation also exhibits low computational  cost because it need not
solve  equivalent   intermediate  \mdp{}s;  the  computation   of  the
bisimulation metric over space $S$  occurs once regardless of how many
times      the      metric      is     used      as      shown      in
Fig.~\ref{fig:stateequivalence_bisimulation}.   \textsc{mwal} requires
$\mathcal{O}(\ln  k)$  ($k$  is  number of  features)  iterations  for
convergence, which  is lower than  the $\mathcal{O}(k \ln k)$  for the
\textsc{projection} method.  Although an iteration of \textsc{firl} is
slower  than  both \textsc{mmp}  and  \textsc{projection}  due to  the
computationally expensive step  of regression, \textsc{firl} converges
in fewer iterations as compared to the latter two methods.

Some  optimization  methods  employ   more  affordable  techniques  of
gradient  computations.  In  contrast with  the fixed-point  method in
\textsc{hybrid-irl},  the approximation  method in  \textsc{birl} with
active learning (reviewed in Section~\ref{subsec:bayesian}) has a cost
that is polynomial in the number of states.  For maximum entropy based
parameter      estimation,      gradient-based     methods      (e.g.,
BFGS~\cite{fletcher1987practical})   outperform    iterative   scaling
approaches~\cite{Malouf_ComparisonEstimatns}.

\textsc{birl} with  active learning offers a  benefit over traditional
\textsc{birl}  by  exhibiting  reduced  sample  complexity.   This  is
because  it seeks  to ascertain  the most  informative states  where a
demonstration is needed, and queries for it.
Consequently, less  demonstrations are  needed and the  method becomes
more targeted. Of course, this  efficiency exists at the computational
expense  of interacting  with the  expert.  Model-free  \textsc{reirl}
uses  fewer samples  (input trajectories)  as compared  to alternative
methods       including       a      model-free       variant       of
\textsc{mmp}~\cite{Boularias2011relative}.

\subsubsection{Continuous State Spaces}

While most approaches for \irl{} target discrete state spaces, a group
of prominent methods that operate  on continuous state spaces are path
integral based approaches, \textsc{pi-irl}.
These  aim  for  local  optimality  of  demonstrations  to  avoid  the
complexity  of full  forward  learning in  a  continuous space.   This
approximation  makes  it  scale well  to  high-dimensional  continuous
spaces and large demonstration data.  Although the performance of path
integral  algorithms is  sensitive to  the  choice of  samples in  the
demonstration, they show promising progress in
scalability.  
Kretzschmar                   et                  al.
  \cite{Kretzschmar_interactingpedestrians}  apply  \textsc{maxentirl}
  to learn  the probability distribution over  navigation trajectories
  of interacting pedestrians using a  subset of their continuous space
  trajectories.  A  mixture distribution models both,  the discrete as
  well as the continuous navigation decisions.

	
\subsubsection{High Dimensional and Large Spaces} 

In  \irl{}  methods  such  as  {\sc maxentirl}  and  {\sc  birl},  the
complexity of computing  the partition function $Z$,  which appears in
the      normalization      constant       of      the      likelihood
(Eq.~\ref{eqn:maxent-likelihood}),  increases  exponentially with  the
dimensionality  of the  state space  because it  requires finding  the
complete policy under the current solution $\hat{R}_E$.
Approaches for making the likelihood computation in a high-dimensional
state  space  tractable include  the  use  of importance  sampling  as
utilized by  {\sc reirl} and  {\sc gcl}, down-scaling the  state space
using                                                  low-dimensional
features~\cite{Vernaza_highdimension_approximation},      and      the
assumption by {\sc pi-irl} that demonstrations are locally optimal.

For  the optimizations  involved in  maximum entropy  methods, limited
memory variable metric  optimization methods such as  L-BFGS are shown
to  perform better  than  other alternatives  because they  implicitly
approximate   the  likelihood   in   the  vicinity   of  the   current
solution~\cite{Malouf_ComparisonEstimatns} thereby limiting the memory
consumption.

        Instead  of  demonstrating  complete  trajectories  for  large
        tasks,   a   problem   designer   may   decompose   the   task
        hierarchically.   An expert  may then  give demonstrations  at
        different levels  of implementation.   The modularity  of this
        process significantly reduces the complexity of learning.  For
        example, Kolter et al.~\cite{Kolter_2007_HierarchicalAL} apply
        such task  decomposition toward learning  quadruped locomotion
        by scaling \irl{} from low- to high-dimensional spaces.
        Likewise,  Rothkopf  et  al.   \cite{Rothkopf_modular_highDim}
        utilize the independence  between the components of  a task --
        each modeled using a stochastic  reward function of its own --
        to introduce decomposition in \textsc{birl}.

		
We  may  speed up  forward
learning by quickly computing the  values of the intermediate policies
learned in \irl{}.
%
Both {\sc  lpal} and  {\sc lpirl} are  incremental extensions  of {\sc
  mwal} that solve the underlying MDP in {\sc mwal} using the dual and
primal   linear   programs,   respectively.   These   linear   program
formulations makes  solving the \mdp{}  less expensive in  large state
spaces with many basis functions ($\phi$ for $R_E=w^T\phi$).
  Similarly,  \textsc{csi} and  \textsc{scirl}  do not  need to  solve
  \mdp{}s  repeatedly because  they  update the  previous solution  by
  exploiting   the  structure   imposed   on  the   \mdp{}  by   their
  classification-based models.
%


\section{Extensions of Basic \irl{}}
\label{sec:Extensions}

Having   surveyed    the   foundational   methods   for    \irl{}   in
Section~\ref{sec:PrimaryMethods}  and discussed  how  they  and a  few
extensions      mitigate      the      various      challenges      in
Section~\ref{sec:Mitigation}, we  now discuss important ways  in which
the  assumptions of  the basic  \irl{}  problem have  been relaxed  to
enable advances toward real-world applications.

	
\subsection{Incomplete and Imperfect Observations} 

Learners in  the real world must  deal with noisy sensors  and may not
perceive the full demonstration  trajectory.  For example, the merging
car  B  in  our   illustrative  example  of  Fig.~\ref{fig:car_merger}
described in Section~\ref{sec:intro} may not  see car A in the merging
lane until it comes into its sensor view. This is often complicated by
the fact that car B's sensor may be partially blocked by other cars in
front      of      it,       which      further      occludes      car
A. Additionally, the expert  itself may possess noisy
  sensors and may not observe its own state perfectly.

\subsubsection{Extended Definition}

The  property of  incomplete  and noisy  observations  by the  learner
modifies  the  traditional  \irl{}  problem   and  we  provide  a  new
definition below for completeness.

\begin{defn}[\irl{}  with  imperfect  perception]  
  Let $\M\backslash_{R_E}$  represent the dynamics of  the expert $E$.
  Let     the     set     of     demonstrated     trajectories     be,
  $\mathcal{D}=\{\langle(s_{0},a_{0}),(s_{1},a_{1}),\ldots
  (s_{j},a_{j})\rangle_{i=1}^N\}$,          $s_{j}\in         Obs(S)$,
  $a_{j}\in Obs(A)$, $i,j,N  \in\mathbb{N}$.  Either some state-action
  pairs of a  trajectory, $\tau \in \mathcal{D}$, are  not observed or
  some of the observed state-action  pairs could be different from the
  actual demonstrated ones.
  Thus, let $Obs(S)$ and $Obs(A)$ be the subsets of states and actions
  respectively that  are observed.   Then, determine  $\hat{R}_E$ that
  best  explains  either  given  policy $\pi_E$  or  the  demonstrated
  trajectories.
  \label{def:extension_imperfect_perception}
\end{defn}


\begin{figure}[ht!]
  \centering
  \includegraphics[scale=.35]{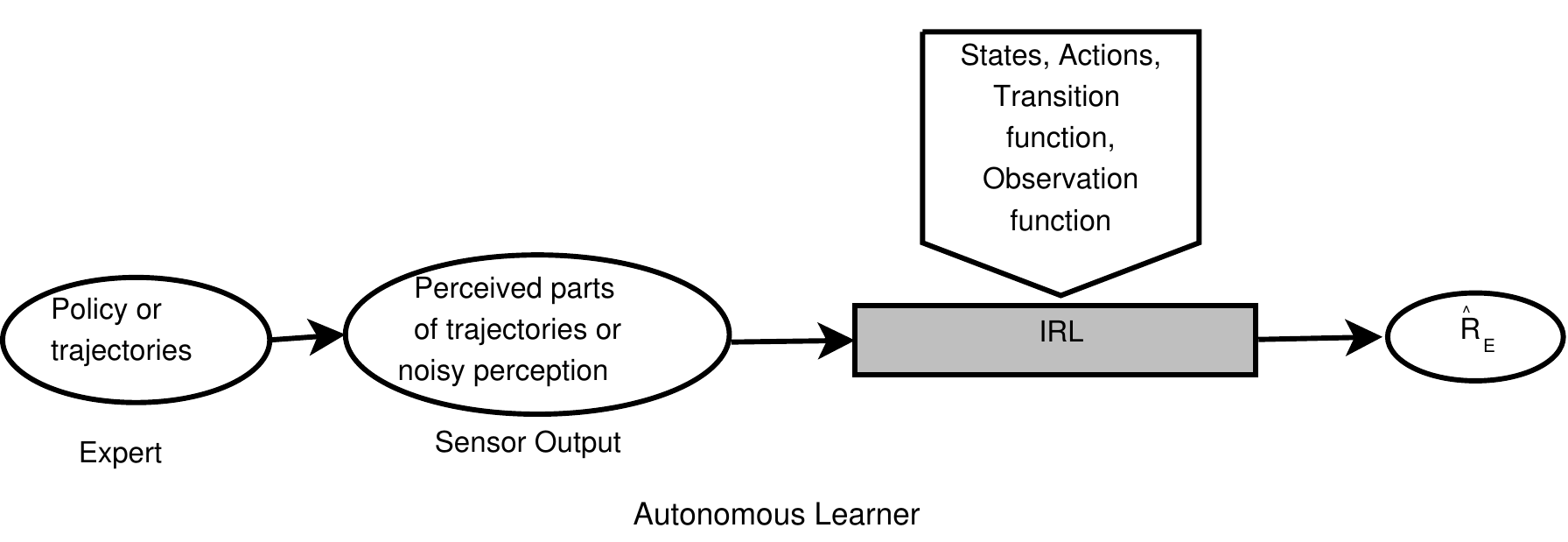}
  \caption{\irl{} with  imperfect perception of the  input trajectory.
    The learner is limited to using just the perceived portions.}
  \label{fig:IRL_incomp_perception}
\end{figure}

Figure~\ref{fig:IRL_incomp_perception} revises  the schematic  for the
traditional \irl{} shown in Fig.~\ref{fig:IRL} to allow for incomplete
and  imperfect observations.   Observing the  trajectories imperfectly
may  require  the learner  to  draw  inferences about  the  unobserved
state-action pairs or the true  ones from available information, which
is challenging.

\subsubsection{Methods}

\begin{figure}[ht!]
\centerline{ 
    \includegraphics[width=0.3\textwidth,height=4cm]{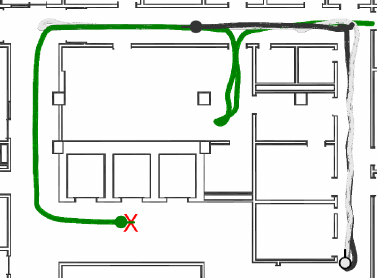}
    \includegraphics[width=0.6\textwidth,height=4cm]{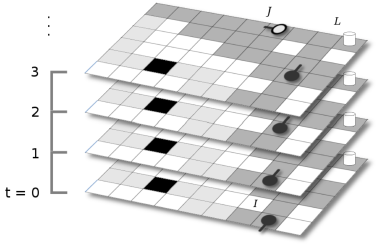}}
  \caption{Prediction   of   experts'  behaviors   using   multi-robot
    \irl{}*~\cite{Bogert_mIRL_Int_2014}  in  a multi-robot  patrolling
    problem  (left).   Learner $L$  (green)  needs  to cross  hallways
    patrolled by experts $I$ (black,  reward $R_{E_1}$) and $J$ (gray,
    reward  $R_{E_2}$).   It  has  to reach  goal  `X'  without  being
    detected.  Due to  occlusion, just portions of  the trajectory are
    visible  to  L.   After  learning  $R_{E_1}$  and  $R_{E_2}$,  $L$
    computes their policies and projects their trajectories forward in
    time and  space to  know the possible  locations of  patrollers at
    each future time  step.  These projections over  future time steps
    help create $L$'s own policy as shown in the figure on the right.}
    \label{fig:occlusion}
\end{figure}

Bogert  et al.~\cite{Bogert_mIRL_Int_2014}  introduce  {\sc irl*}  for
settings where the learner is unable to see some state-action pairs of
the demonstrated  trajectories due to occlusion.   The maximum entropy
formulation  of  the  structured  apprenticeship  learning  method  by
Boularias et  al. is  generalized to  allow feature  expectations that
span the observable state space only.  This method is applied to a new
domain    of     multi-robot    patrolling    as     illustrated    in
Fig.~\ref{fig:occlusion}.

The  principle  of   latent  maximum  entropy~\cite{Wang2002,Wang2012}
allows us  to extend  the maximum entropy  principle to  problems with
hidden   variables.     By   using    this   extension,    Bogert   et
al.~\cite{Bogert_EM_hiddendata_fruit}  continued  along  the  vein  of
incomplete observations and generalized {\sc maxentirl} to the context
where a dimension of the expert's actions are hidden from the learner.
For example, the amount of force applied by a human while picking ripe
and unripe  fruit usually  differs but  this would  be hidden  from an
observing  co-worker  robot.   An expectation-maximization  scheme  is
introduced  with the  E-step involving  an expectation  of the  hidden
variables while the M-step performs the {\sc maxent} optimization.

\begin{figure}[ht!] 
  \centering{
    \includegraphics[scale=0.55]{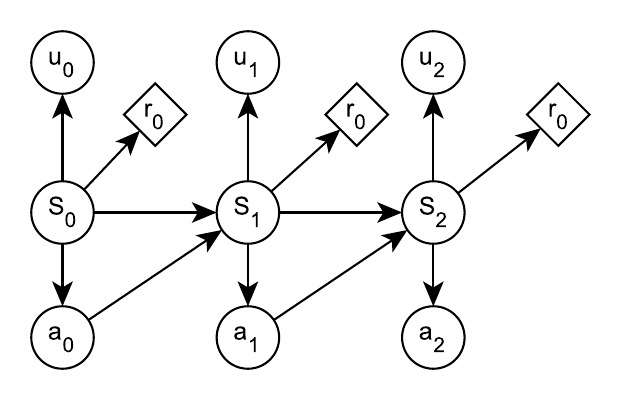}
    \caption{Hidden-variable \mdp{}.  $u_i$ is a noisy observation, by
      the  learner, of  the state  $s_i$ reached  after taking  action
      $a_{i-1}$.       The     source      of     illustration      is
      \cite{hiddenMDP_Kitani2012}.  The  figure  is  shown  here  with
      permission from publisher.}\label{fig:hMDP}
  }
\end{figure} 
	
Taking  the context  of noisy  observations, a  hidden-variable \mdp{}
incorporates   the   probability   of  learner's   noisy   observation
conditioned on the  current state ($u$ in  Fig.~\ref{fig:hMDP}), as an
additional   feature   $\phi_u$   in  the   feature   vector   $\phi$.
Hidden-variable           inverse            optimal           control
(\textsc{hioc})~\cite{hiddenMDP_Kitani2012}        then       modifies
\textsc{maxentirl} to a problem where  the dynamics are modeled by the
hidden variable \mdp{} with a linearly-weighted reward
function. 
Consequently, the  expression for the likelihood  of expert's behavior
incorporates  the additional  feature and  its weight  ($\phi_u,w_u$).
The  tuning  of weights  during  optimization  also adjusts  $w_u$  to
determine the reliability of the imperfect
observations.  
	

\begin{figure}[ht!] 
  \centering{
    \includegraphics[scale=1]{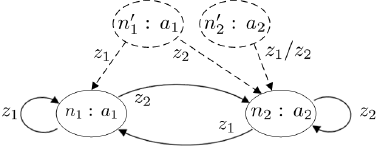}
    \caption{In  this illustration,  similar  to the  one  in Choi  et
      al.~\cite{Choi2011},  consider a \pomdp{}  with two  actions and
      two observations.  $\pi_E$ (solid lines) is  a \textsc{fsm} with
      nodes   $\{n_1,n_2\}$  associated  to   actions  and   edges  as
      observations  $\{z_1,z_2\}$.   The  one-step deviating  policies
      (dashed  lines)  $\{\pi_i\}_{i=1}^2$   are  policies  which  are
      slightly  modified  from $\pi_E$.   Each  $\pi_i$ visits  $n_i'$
      instead  of  $n_i$  and  then  becomes  same  as  $\pi_E$.   The
      comparison    of    $\hat{V}^{\pi_E}$   with    $\{V(\pi_i)\}_{i=1}^2$
      characterizes  the  set  of  potential  solutions.   Since  such
      policies  are  suboptimal  yet  similar to  expert's, to reduces 
      computations, they  are
      preferable for comparison instead of comparing $\hat{V}^{\pi_E}$ 
      with all possible
      policies.}
    \label{pomdp_irl_characterization}
  }
\end{figure}

Choi and Kim~\cite{Choi2011} take a different perspective involving an
expert that senses  its state imperfectly. The expert is  modeled as a
partially  observable  \mdp{} (\pomdp{})~\cite{Kaelbling_pomdp}.   The
expert's uncertainty about its current  physical state is modeled as a
belief (distribution)  over its state  space.  The expert's  policy is
then a mapping  from its beliefs to optimal actions.   The method {\sc
  pomdp-irl} makes either this policy  available to the learner or the
prior  belief along  with the  sequence of  expert's observations  and
actions  (that can  be used  to reconstruct  the expert's  sequence of
beliefs).   The  \pomdp{}  policy  is represented  as  a  finite-state
machine  whose  nodes   are  the  actions  to   perform  on  receiving
observations that form the edge labels.
The learner conducts a search through the space of reward functions as
it gradually improves on the previous policy until the policy explains
the   observed    behavior.    Figure~\ref{pomdp_irl_characterization}
illustrates this approach.
However, a  well-known limitation of  utilizing \pomdp{}s is  that the
exact  solution  of  a  \pomdp{}  is  PSPACE-hard,  which  makes  them
difficult to scale to pragmatic settings.

In   {\sc  pomdp-irl},   the  expert   may  not   observe  its   state
perfectly. However, \irl{}* and  \textsc{HIOC} differ from this setup.
They  model  the  learner  observing the  expert's  state  and  action
imperfectly whereas the expert is perfectly aware of its state.

\subsection{Multiple Tasks} 

Human drivers often exhibit differing  driving styles based on traffic
conditions as  they drive toward  their destination. For  example, the
style of  driving on  the rightmost  lane of  a freeway  is distinctly
different prior  to the joining of  a merging lane, at  joining of the
lane, and post joining the lane.  We may model such distinct behaviors
of expert(s)  as guided  by differing reward  functions. Consequently,
there  is a  need to  investigate methods  that learn  multiple reward
functions simultaneously.

\subsubsection{Extended Definition}

To accommodate demonstrations involving multiple tasks, we revise the
traditional \irl{} problem definition as given below.         

\begin{defn}[Multi-task \irl{}]  Let the dynamics of  the expert(s) be
  represented by  $M$ \mdp{}s each  without the reward  function where
  $M$ may not  be known. Let the set of  demonstrated trajectories be,
  $\mathcal{D}=\{\langle(s_{0},a_{0}),(s_{1},a_{1}),\ldots,
  (s_{j},a_{j})\rangle_{i=1}^{N}\}$,   $s_{j}\in  S$,   $a_{j}\in  A$,
  $i,j,N \in\mathbb{N}$.
  Determine $\hat{R}_E^1$, $\hat{R}_E^2$, $\ldots$, $\hat{R}_E^M$ that
  best explain the observed behavior.
  \label{def:extension_multi_task_irl}
\end{defn}
	
\begin{figure}[ht!]
  \centering
  \includegraphics[scale=.35]{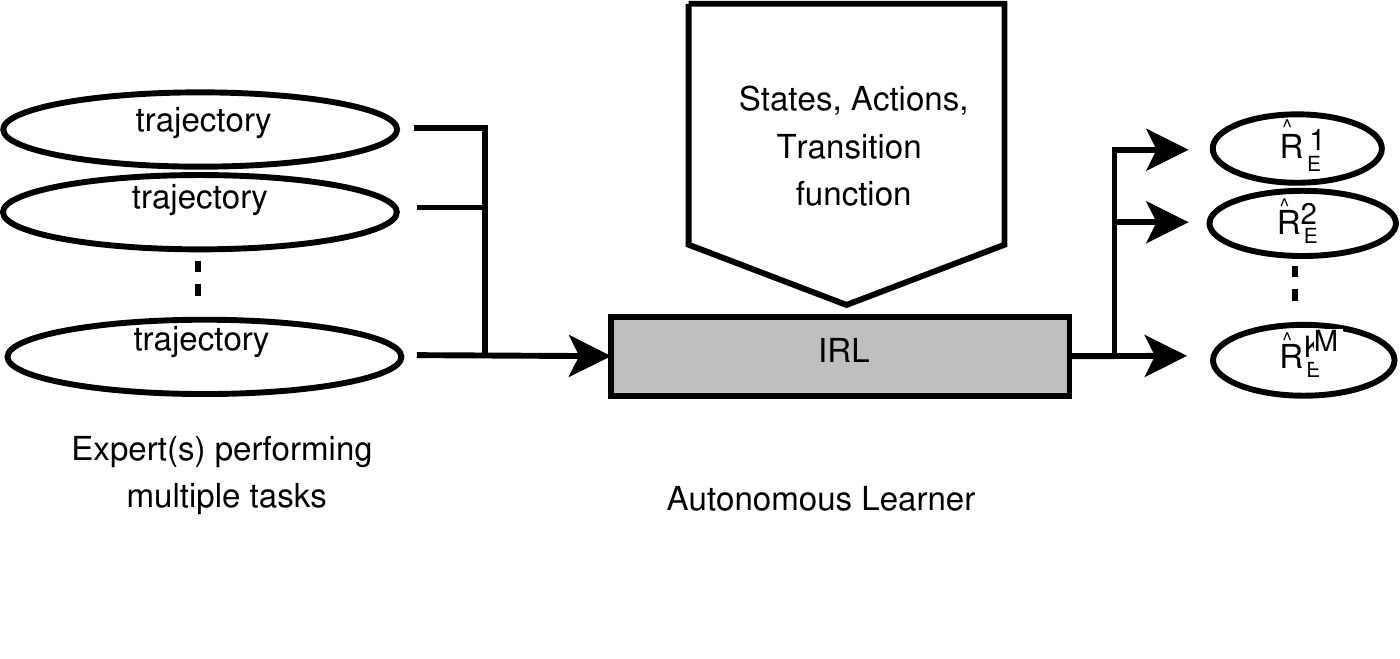}
  \caption{Multi-task \irl{}  involves learning multiple
    reward functions.   The input is  mixed trajectories
    executed by a single or multiple expert(s) realizing
    behavior driven by  different reward functions.  The
    output  is  a  set  of reward  functions,  each  one
    associated  with  a  subset  of  input  trajectories
    potentially generated by it.}
  \label{fig:IRL_multipletasks}
\end{figure}
	

Figure~\ref{fig:IRL_multipletasks}  gives   the  schematic   for  this
important \irl{}  extension.  Having  to associate  a subset  of input
trajectories from the  demonstration to a reward  function that likely
generates it  (also called  the data  association problem)  makes this
extension challenging.   This becomes further complex  when the number
of involved tasks is not known.
Diverse  methods  have  sought  to  address  the  problem  defined  in
Def.~\ref{def:extension_multi_task_irl},
and we briefly review them below.

\subsubsection{Methods}

Babes-Vroman  et  al.~\cite{Babes-Vroman2011}  assume  that  a  linear
reward function of an expert can change over time in a chain of tasks.
The method aims to  learn multiple reward functions with common
features $\{\hat{R}_E^i=\bm{w}_i^T\phi\}_{i=1}^M, M \in  \mathbb{N}$. 
Given prior knowledge of $M$, the  solution is a pair of weight vector
$\bm{w}_i\in \mathbb{R}^k$  and a correspondence probability  for each
reward function $\hat{R}_E^i$.  This  probabilistically ties a cluster
of  trajectories  to  a  reward  function.   The  process  iteratively
clusters  trajectories  based  on   current  hypothesis,  followed  by
implementation  of  \textsc{mlirl}  for updating  the  weights.   This
approach is reminiscent of using expectation-maximization for Gaussian
data clustering.

	
\begin{figure}[ht!] 
  \centering{
  	\includegraphics[width=\textwidth]{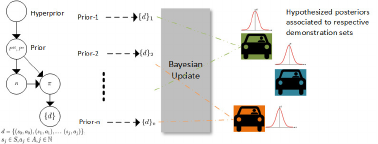}
    \caption{Parametric   multi-task  \textsc{birl}.
      Tuning  of  independent  hyperprior modifies  the  dependent
      priors $P^R$  on reward functions  and priors $P^\pi$ on  policies, the
      diagram  shows it  as a  sequence  of priors.  The variation  in
      priors  is  assumed   to  influence  the  observed  trajectories
      $\{\tau\}_i,i\in  \mathbb{N}$. Bayesian  update outputs an
      approximated  posterior over  rewards and  policies.
    }\label{fig:Multi_Task_BIRL}
  }
\end{figure} 

Continuing  with the  assumption of  knowing  $M$, {\sc  birl} can  be
generalized  to   a  hierarchical  Bayes  network   by  introducing  a
hyperprior that imposes  a probability measure on the  space of priors
over    the    joint    reward-policy   space.     Dimitrakakis    and
Rothkopf~\cite{Dimitrakakis2012} show how the prior is sampled from an
updated posterior  given an input demonstration.   This posterior (and
thus the sampled prior) may  differ for an expert performing different
tasks or multiple experts involved in different
tasks. 
Within      the     context      of      our     running      example,
Fig.~\ref{fig:Multi_Task_BIRL}  illustrates how  this approach  may be
used to learn  posteriors for multiple drivers on a  merging lane of a
freeway.



In contrast to parametric  clustering, \textsc{dpm}-\textsc{birl} is a
clustering  method that  learns  multiple  reward specifications  from
unlabeled fixed-length trajectories~\cite{Choi2012}.   It differs from
the previous methods in this section in that the number of experts are
not    known.     Therefore,    it   addresses    the    problem    in
Def.~\ref{def:extension_multi_task_irl} with $M$  unknown.  The method
initializes  a  nonparametric Dirichlet  process  of  priors over  the
reward functions and aims to assign  each input trajectory to a reward
function that  potentially generates  it, thereby forming  clusters of
trajectories.  Learning  occurs by  implementing a Bayesian  update to
compute  the  joint  posterior  over  the  reward  functions  and  the
probabilistic assignments  to clusters.  The procedure  iterates until
the reward functions and clusters stabilize.

In   settings   populated  by   multiple   experts,   Bogert  et   al.
\cite{Bogert_mIRL_Int_2014} extend  {\sc irl*} and {\sc  maxentirl} to
multiple experts who may interact, albeit sparsely.  These experts are
mobile robots patrolling  a narrow hallway. While  the motion dynamics
of each expert is modeled separately,  the interaction is modeled as a
strategic  game  between  the  two  experts;  this  approach  promotes
scalability to many  experts.  Experts are assumed to play  one of the
Nash equilibria profiles during  the interaction, although the precise
one is unknown to the learner.
Alternatively,  all experts  may be  modeled jointly  as a  multiagent
system.  Reddy  et al.~\cite{Reddy_DecMulti}  adopt this  approach and
model  multiple interacting  experts  as  a decentralized  general-sum
stochastic game.   Lin et al. \cite{Lin_MultiGame}  propose a Bayesian
method learning the distribution over rewards in a sequential zero-sum
stochastic multi-agent game.

\subsection{Incomplete Model} 

Definition~\ref{def:irl}  for \irl{}  assumes  full  knowledge of  the
transition  model  $T$ and  the  reward  feature functions.   However,
knowing the transition probabilities that represent the dynamics
or  specifying  the complete  feature  set  is challenging  and  often
unrealistic.   Hand-designed  features  introduce  structure  to  the
reward, but they increase the engineering
burden.  
Inverse learning is difficult when the learner is partially unaware of
the expert's dynamics  or when the known features  do not sufficiently
model  the  expert's  preferences.   Subsequently,  the  learner  must
estimate the missing components for inferring
$\hat{R}_E$.  
Readers  familiar with  \rl{} may  notice that  these
  extensions  share   similarity  with  model-free  \rl{}   where  the
  transition model and reward function features are also unknown.

\subsubsection{Extended Definition}


\begin{figure}[ht!]
  \centering
  \includegraphics[scale=.4]{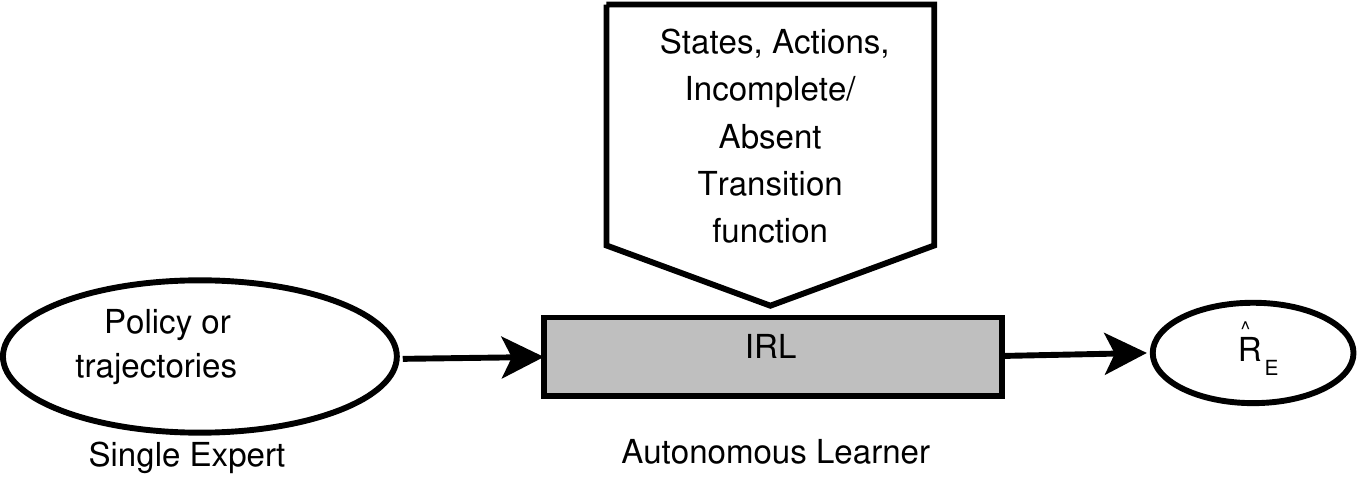}
  \caption{\irl{} with incomplete model of transition probabilities.}
  \label{fig:IRL_incompleteprobabilities}
\end{figure}

\begin{defn}[Incomplete  dynamics]  Let   an  \mdp{}  without  reward,
  $\mathcal{M}\backslash_{R_E}=(S,A,\hat{T},\gamma)$,   represent  the
  dynamics  between an  expert  and its  environment, where  $\hat{T}$
  specifies   the  probabilities   for  a   subset  of   all  possible
  transitions.         The        input        is        demonstration
  $\mathcal{D}            =\{\langle(s_{0},a_{0}),(s_{1},a_{1}),\ldots
  (s_{j},a_{j})\rangle_{i=1}^N\}$,   $s_{j}\in   S$,   $a_{j}\in   A$,
  $i,j,N \in\mathbb{N}$  or expert's policy $\pi_E$.   Then, determine
  reward  $\hat{R}_E$  that  best  explains either  the  input  policy
  $\pi_E$ or the observed demonstration $\mathcal{D}$.
  \label{def:extension_incomplete_dynamics}
\end{defn} 

Figure~\ref{fig:IRL_incompleteprobabilities}
  illustrates the corresponding generalized \irl{} pipeline.  Next, we
  define the \irl{} problem when the set of basis feature functions is
  incomplete.

\begin{defn}(Incomplete  features)  Let   an  \mdp{}  without  reward,
  $\mathcal{M}\backslash_{R_E}=(S,A,T,\gamma)$, represent the dynamics
  of  an  expert  and  its   environment.   Let  the  reward  function
  $\hat{R}_E=f(\bm{\phi})$ depend on the feature set $\bm{\phi}$.  The
  input                        is                        demonstration
  $\mathcal{D}
  =\{\langle(s_{0}^{i},a_{0}^{i}),(s_{1}^{i},a_{1}^{i}),\ldots
  (s_{j}^{i},a_{j}^{i})\rangle_{i=1}^N\}$, $s_{j}\in S$, $a_{j}\in A$,
  $i,j,N  \in\mathbb{N}$ or  expert's  policy $\pi_E$.   If the  given
  feature  set  $\bm{\phi}$  is  incomplete,  find  the  features  and
  function $\hat{R}_E$ that best explains the input.
  \label{def:extension_incomplete_features}
\end{defn}

\subsubsection{Methods}

While  the  majority of  \irl{}  methods  assume completely  specified
dynamics, we briefly review two that learn the dynamics in addition to
the  reward function.   \textsc{mwal} obtains  the maximum  likelihood
estimate  of   unknown  transition  probabilities  by   computing  the
frequencies  of state-action  pairs  which are  observed  more than  a
preset threshold number of times.  The process determines the complete
transition  function  by routing  the  transitions  for the  remaining
state-action pairs to  an absorbing state.  To  formally guarantee the
accuracy  of learned  dynamics and  thereby the  reward function,  the
algorithm leverages a  theoretical upper bound on the  accuracy of the
learned  transition model  if  the learner  receives  a given  minimum
amount of
demonstration~\cite{Syed2008_supplement}.  

\begin{figure}[ht!] 
  \centering{ \includegraphics[scale=.75]{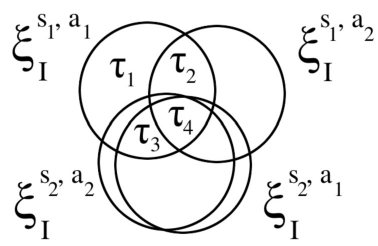}}
    \caption{In           m\irl{}*\textbackslash           \textsc{t},
      $\xi^{(s,a)}_i=\{\tau_1,\tau_2,\ldots\tau_b\}$    denotes    the
      transition-features  for transition  $(s,a,s')$  of expert  $i$,
      $s'$   is  intended   next   state.    Computation  of   unknown
      probabilities  by  using probabilities  of  transition-features,
      $\prod_{\tau\in\xi^{(s,a)}_i}P(\tau)=T_{sa}(s,a,s')$,         is
      feasible because different transitions share transition features
      among      them.       Source       of      illustration      is
      \cite{Bogert_mIRL_woT_Int_2015}  and  figure is  reprinted  with
      author's permission.}
  \label{fig:transition-features}
\end{figure}

While \textsc{mwal} assumes that a  learner fully observes the states,
m\irl{}*$_{\backslash   T}$~\cite{Bogert_mIRL_Int_2014}   focuses   on
limited  observations   with  unknown  transition   probabilities  and
multiple experts. Bogert  and Doshi model each transition  as an event
composed of  underlying components. For  example, movement by  a robot
may  be decomposed  into  its left  and right  wheels  moving at  some
angular  velocity. Therefore,  the  probability  of reaching  intended
location by moving forward is the  joint probability of left and right
wheels rotating  with the same  velocities. The learner is  assumed to
know the intended next state  for a state-action pair, and probability
not assigned  to the intended  state is distributed equally  among the
unintended  next states.   Importantly,  the  components, also  called
transition features,  are likely to  be shared between  observable and
unobserved transitions as shown in Fig.~\ref{fig:transition-features}.
Therefore,  a   fixed  distribution   over  the   transition  features
determines  $T$.   The  frequencies  of a  state-action  pair  in  the
demonstration  provide  a set  of  empirical  joint probabilities,  as
potential  solutions. The  preferred solution  is the  distribution of
component     probabilities     with      the     maximum     entropy.
m\irl{}*\textsc{\textbackslash    t}     generalizes    better    than
\textsc{mwal} because  the structure introduced by  shared features is
more generalizable  in the  space of  transition probabilities  than a
local frequency based estimation.
	
On the other hand, Boularias et al. \cite{Boularias2011relative} shows
that  the   transition  models  approximated  from  a   small  set  of
demonstrations may result in  highly inaccurate solutions. 
To  this end,  \textsc{pi-irl} learns  a reward  function without  any
input  transition  probabilities.   Furthermore,  for  estimating the 
unknown dynamics, \textsc{gcl}~\cite{Levine_UnknownDynamics_NN_policy}
iteratively  runs a  linear-Gaussian  controller  (current policy)  to
generate trajectory  samples, fits  local linear-Gaussian  dynamics to
them by using linear regression,  and updates the controller under the
fitted dynamics.
		
A  generalization of  \textsc{mmp}  that focuses  on  \irl{} when  the
feature vector  is known  to be insufficient  to explain  the expert's
behavior is  \textsc{mmpboost}~\cite{Ratliff2007}.  In this  case, the
method assumes that a predefined  set of primitive features, which are
easier to specify, create the  reward feature functions.  In the space
of nonlinear functions  of base features, {\sc  mmpboost} searches new
features that make  the demonstrated trajectories more  likely and any
alternative (simulated)  trajectories less likely.   Consequently, the
hypothesized  reward  function  performs  better  than  the  one  with
original feature  functions.  Further, it  is well known  that methods
employing L1 regularization objectives  can learn robustly when input
features               are                not               completely
relevant~\cite{Ng_RegularizationComparison}. 
In addition to  {\sc mmpboost}, \textsc{gpirl} also  uses this concept
of  base features  to learn  a new  set of  features which  correspond
better     to     the     observed     behavior. 

	
In  some applications,  it is  important to  capture the  {\em logical
  relationships} between  base features  to learn an  optimum function
representing the expert's reward.  Most methods do not determine these
relationships  automatically.  Recall  that  \textsc{firl}  constructs
features by  capturing these  relationships in  a regression  tree. In
contrast, \textsc{bnp}-\textsc{firl} uses an  Indian buffet process to
derive a Markov Chain Monte  Carlo procedure for Bayesian inference of
the     features     and     weights    of     a     linear     reward
function~\cite{Choi2013bayesian}.   \textsc{bnp-firl} is  demonstrated
to construct features  more succinct than those  by \textsc{firl}.  Of
course, all  these methods  are applicable only  in domains  where the
feature  space  is sufficient  to  satisfactorily  express the  reward
function.
	
\subsection{Nonlinear Reward Function} 

A  majority  of  the   \irl{}  methods  such  as  \textsc{projection},
\textsc{mmp},  and \textsc{maxentirl}  assume that  the solution  is a
weighted   linear   combination   of   a  set   of   reward   features
(Eq.~\ref{eqn:linear-reward}).   While  this  is sufficient  for  many
domains, a  linear representation  may be  over simplistic  in complex
real tasks  especially when raw sensory  input is used to  compute the
reward   values~\cite{Finn_gcl}.    Also,  analyzing   the   learner's
performance w.r.t.  the best solution  seems compromised when a linear
form restricts  the class  of possible  solutions.  But  a significant
challenge  for  relaxing  this  assumption is  that  nonlinear  reward
functions may take any shape, which  could lead to a very large number
of parameters and search space.

As  our definition  of  \irl{} given  in  Def.~\ref{def:irl} does  not
involve the structure of the  learned reward function, it continues to
represent the problem in the  context of nonlinear reward functions as
well.

\begin{figure}[ht!] 
  \centering{
  	\includegraphics[scale=1]{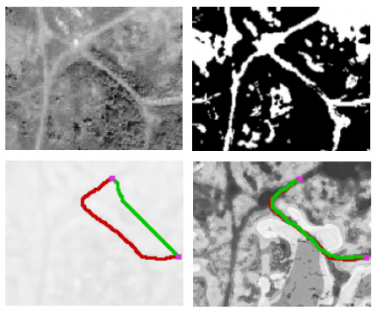}
    \caption{  Learning  a  nonlinear  reward function  with  boosted
      features improves performance  over linear reward. Learner needs
      to imitate  example paths drawn  by humans in  overhead imagery.
      Upper  left panel  - base  features for  a region.   Upper right
      panel -  image of  the region  used for testing.  Red path  is a
      demonstrated path and  Green path is a learned  path. Lower left
      panel - a map (un-boosted linear reward function) inferred using
      \textsc{mmp} with  uniformly high cost  everywhere.  Lower right
      panel   shows    results   of   \textsc{mmp}\textsc{boost}.    Since
      \textsc{mmp}\textsc{boost}  creates  new  features by  a  search
      through  a  space of  nonlinear  reward  functions, it  performs
      significantly   better.   We    reprinted   this   figure   from
      \cite{Ratliff2007} with permission from MIT press.}}
\label{Boosted_features_vs_unboosted_linear}
\end{figure} 

To overcome the constraint of  using a linear reward function, methods
\textsc{mmpboost},   \textsc{learch},  and   \textsc{gpirl}  infer   a
nonlinear  reward function.   {\sc mmpboost}  and {\sc  learch} use  a
matrix  of features  in an  image (cost  map) and  {\sc gpirl}  uses a
Gaussian   process    for   representing   the    nonlinear   function
$R_E=f(\phi)$. Figure~19 
shows  the  benefit of  a  nonlinear  form  with boosted  features  as
compared to a restrictive linear form. In addition to these, Wulfmeier
et al.~\cite{Wulfmeier2015} and  Finn et al.~\cite{Finn_gcl} represent
a   complex   nonlinear  cost   function   using   a  neural   network
approximation, thereby avoiding the assumption of a linear form.\\

\noindent Table    ~\ref{tbl:challengesAddressedByExtensions}   abstracts    and
summarizes  the  key  properties  of  the  methods  reviewed  in  this
section.  Some of  these  methods build  on  the foundational  methods
reviewed  in  Section~\ref{sec:PrimaryMethods}  while others  are  new
introduced with the aim of generalizing \irl{} in pragmatic ways.
	



\begin{table}[h!]
	\begin{center}
		{\scriptsize
			\bgroup
			\def\arraystretch{1.2}
			\begin{tabular}{ | l || c | c| l |}
			\hline
			\textbf{Method} &
			\textbf{$\hat{R}_E$ params} &
			\textbf{Optimization Obj.} & 
			\textbf{Notable aspects} \\ \hline 
                          \multicolumn{4}{|l|}{\bf Methods for
                         incomplete and noisy observations}\\ \hline
			{\sc  pomdp-irl} & \multirow{4}{*}{$\bm{w}$} &
                                                                       feature
                                                  expectation of
                                                                       policy
                                                   &
                                                     \textsc{max-margin} with noisy \\ 
			&   &                                                                        -
                                                   empirical feature
                                                                       exp.   &  observations of expert \\
			\cline{1-1}\cline{3-4}
   			\textsc{hioc}  &  &  entropy of distribution  &
             modeling noise in input  \\ 
                          \cline{1-1}\cline{4-4}
			\textsc{irl$^*$} &   &  over trajectories  &
                                                                    \irl{} with hidden variables\\  
                                                                    \hline
            \multicolumn{4}{|l|}{\bf Methods for multiple tasks}\\ \hline
			{Dimitrakakis et al. \cite{Dimitrakakis2012}}
                                        &  $
                                          \{R_E^i\}^M_{i=1}  $ &
                                                                       joint
                                                                       dist.
                                                   over & hierarchical \textsc{birl} for  \\
			& & rewards and policies & multiple hypotheses \\ \hline		
			\textsc{dpm-birl} & $ \{R_E^i\}^* $ & generative DP & first nonparametric\\
			& & governing $\mathcal{D}$ & multi-task technique \\ \hline 
			{Reddy et al. \cite{Reddy_DecMulti}} & $
                                                               \{R_E^i\}^M_{i=1}
                                                               $ &
                                                                   joint
                                                                   policy
                                                                   value
                                                   &  modeling expert interactions \\ 
			{Lin et al. \cite{Lin_MultiGame}}& & & using
                                                               game theory \\ \hline
            \multicolumn{4}{|l|}{\bf Methods for
            incomplete model parameters}\\ \hline		
			\textsc{mmpboost} & \multirow{4}{*}{$ \bm{w}
                                            $} &  value of observed
                                                 $\tau$ & max. likelihood  derived  \\ 
			& & - max of values from all other $\tau$ &  classifier to fit $\phi$ \\ 
			\cline{1-1}\cline{3-4}
            \textsc{mwal} &  & min diff. in value of policy & first formal bound on \\ 
			& & and observed $\tau$ across features &  learning dynamics \\ \cline{1-2}\cline{3-4}
			\textsc{bnp-firl} & $\bm{w}$, primitive & generative IBP governing & integrating feature  \\ 
			& features & $\{\text{primitive features},\bm{w},\mathcal{D}\}$ & learning in \textsc{birl} \\ \hline
		\end{tabular}
		\caption{A comparative analysis of the challenges addressed by the extensions introduced in Section~\ref{sec:Extensions}. Please refer to Table~\ref{tbl:notations1} for the explanation of abbreviations and notations used here.}
		\label{tbl:challengesAddressedByExtensions}
		\egroup
	}
	\end{center}
\end{table}

\section{Concluding Remarks and Future Work}
\label{sec:future_work} 

Since the introduction of \irl{}  in 1998 by Russell, researchers have
demonstrated a  significantly improved  understanding of  the inherent
challenges,  developed  various  methods  for  their  mitigation,  and
investigated  the  extension  of these  challenges  toward  real-world
applications.   This  survey takes  a  rigorous  look at  \irl{},  and
focuses  on  the specific  ways  by  which various  methods  mitigated
challenges and contributed to the ongoing progress of \irl{} research.
The  reason  for  this  focus  is  that  we  believe  that  successful
approaches in \irl{} will eventually  combine the synergy of different
methods to solve complex learning problems that typically exhibit many
challenges. 

Our improved understanding has also revealed more novel questions.  In
our  survey of  several  \irl{}  methods, we  observed  that very  few
methods  provably analyzed  the  sample or  time  complexity of  their
techniques,  and compared  it with  those of  other methods.   Indeed,
\textsc{projection} and  \textsc{mwal} are the only  methods among the
foundational  ones that  provide a  sample complexity  analysis. These
methods use  Hoeffding's inequality to  relate the error  in estimated
feature  expectations with  the  minimum sample  complexity. As  such,
there is a  general lack of theoretical guidance on  the complexity of
\irl{} as  a problem,  and on  the complexity  and accuracy  of \irl{}
methods,  with most  focusing  on empirical  comparisons to  establish
improvement.

A  particularly egregious  shortcoming  is that  the  existing set  of
methods do  not scale reasonably to  beyond a few dozens  of states or
more than  ten possible  actions. This is  a critical  limitation that
limits  \irl{}  demonstrations mainly  to  toy  problems and  prevents
\irl{} from being applied in more pragmatic applications. Many methods
in \irl{} rely on parameter  estimation techniques, and current trends
show  that   meta-heuristic  algorithms   can  estimate   the  optimal
parameters  efficiently.  Some  prominent  meta-heuristic methods  are
cuckoo   search   algorithm~\cite{yang2009cuckoo,yang2010engineering},
particle   swarm  optimization~\cite{eberhart1995particle},   and  the
firefly   algorithm~\cite{yang2010firefly}.    As   their   noticeable
benefits, meta-heuristic  algorithms do  not rely on  the optimization
being convex, rather  they can search general  spaces relatively fast,
and they can find a global minimum.  Thus, studying the performance of
these techniques in \irl{} should reveal new insights.

There  is a distinct  lack of  a standard testbed  of problem
domains  for  evaluating \irl{}  methods,  despite  the prevalence  of
empirical  evaluations in  this  area.  Well  designed testbeds  allow
methods to be  evaluated along various relevant  dimensions, point out
shared  deficiencies,  and  typically  speed   up  the  advance  of  a
particular field.  For example,  UCI's machine learning repository and
OpenAI's Gym  library are playing  significant roles in  advancing the
progress   of  supervised   and  reinforcement   learning  techniques,
respectively.

In addition to these immediate avenues of future work, we also discuss
lines of  inquiry below that could  lead to a better  understanding of
\irl{} and lead to progress over the longer term.


\vspace{-0.1in}
\paragraph{Direct and indirect learning} When the state space is large
and precise  identification of  $\hat{\pi}_E$ is  cumbersome, directly
learning  a reward  function  results in  a  better generalization  as
compared    to   policy    matching   \cite{Neu2007}    (see   Section
~\ref{subsection:labelDirectVsIndirect}).    However,  the   issue  of
choosing between  these two  ways of  learning from  demonstrations or
exploiting their synergies warrants a more thorough analysis.

\vspace{-0.1in}
\paragraph{Heuristics} Choi et  al.~\cite{Choi2011} observes that when
the values of learned policy $\hat{\pi}_E$ and expert's policy $\pi_E$
are evaluated on the learned  reward $\hat{R}_E$, both are optimal and
about equal.   However, $\hat{\pi}_E$ obtained using  $\hat{R}_E$ does
not achieve  the same value as  $\pi_E$ when they use  the true reward
$R_E$ for the evaluation.  This is, of course, a quantification of the
reward  ambiguity challenge,  which  we pointed  out  earlier in  this
survey. It  significantly limits  learning accuracy.  We  believe that
the choice of heuristics in the optimization may mitigate this issue.
		
\begin{figure}[ht!] 
  \centering{
    \includegraphics[scale=0.5]{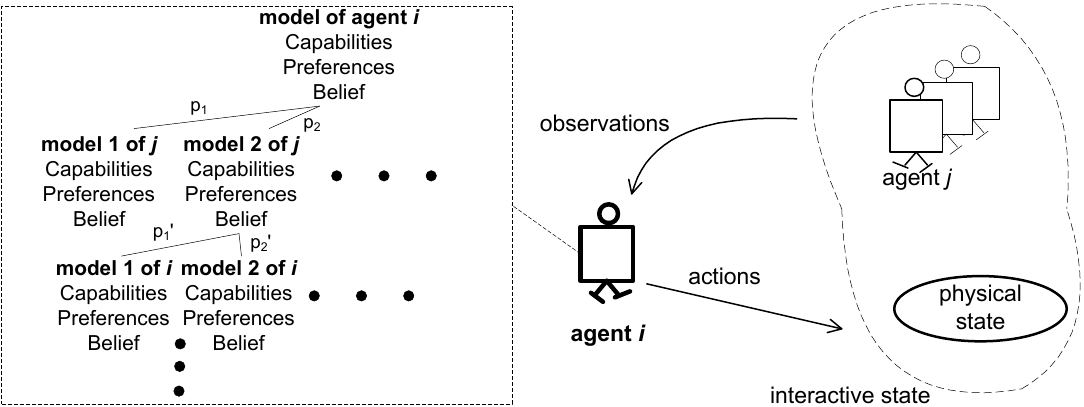}}
    \caption{The   state   of   \textsc{i}-\pomdp{}  evolves   as   an
      \textit{interactive state space} that encompasses the computable
      models (beliefs, preferences, and capabilities) for other agents
      and the physical states of  the environment. Agent $i$ maintains
      and updates his models of other agents.}
\label{fig:I-pomdp}
\end{figure}

\vspace{-0.1in}
\paragraph{Multi-expert  interaction}   Recent  work  on   \irl{}  for
multi-agent  interactive  systems  can  be extended  to  include  more
general classes of interaction-based  models to increase the potential
for  applications~\cite{Lin_MultiGame,Reddy_DecMulti}.  These  classes
include   models    for   fully-observable   state    spaces   (Markov
games~\cite{Littman1994markov_games},   multi-agent  Markov   decision
processes~\cite{Boutilier1999},   interaction-driven    Markov   games
\cite{Spaan_InteractiveFulObs})  and  for partially-observable  states
(partially        observable        identical-payoff        stochastic
games~\cite{Peshkin00},        multi-agent        team        decision
problems~\cite{PynadathT02},  decentralized Markov  decision processes
\cite{Bernstein02},  and I-\pomdp{}s~\cite{Gmytrasiewicz_Doshi_IPOMDP}
illusrated in Fig.~\ref{fig:I-pomdp}).
Outside the domain  of \irl{}, we note  behavior prediction approaches
related  to  inverse  optimal control  in  multi-agent  game-theoretic
settings~\cite{WaughZB_InverseEquilibrium}.       The     regret-based
criterion  in this  work can  be used  for Markov  games too:  for any
linear reward  function, the  learned behavior  of agents  should have
regret less than or equal to that in observed behavior.

\vspace{-0.1in}
\paragraph{Non-stationary rewards} Most methods  assume a fixed reward
function that does  not change.  However, the  preferences of agent(s)
may  change with  time, and  the reward  function can  be time-variant
i.e.,  $R:S\times  A\times  \eta  \to  \mathbb{R}$.   Babes-Vroman  et
al.~\cite{Babes-Vroman2011} capture  such dynamic reward  functions as
multiple reward  functions, but this  approximation is crude.   A more
reasonable start  in this  research direction is  the reward  model in
Kalakrishnan et al.~\cite{Kalakrishnan_continousspace}.

\section{Acknowledgments}
This research  was partly funded by  a grant from the  Toyota Research
Institute of North America and a grant from ONR N-00-0-141310870.

\section{References}
\bibliographystyle{BST/elsarticle-num}
\bibliography{adaij16}

\begin{thebibliography}{10}
\expandafter\ifx\csname url\endcsname\relax
  \def\url#1{\texttt{#1}}\fi
\expandafter\ifx\csname urlprefix\endcsname\relax\def\urlprefix{URL }\fi
\expandafter\ifx\csname href\endcsname\relax
  \def\href#1#2{#2} \def\path#1{#1}\fi

\bibitem{Russell1998}
S.~Russell, Learning agents for uncertain environments (extended abstract), in:
  Proceedings of the Eleventh Annual Conference on Computational Learning
  Theory, COLT' 98, ACM, New York, NY, USA, 1998, pp. 101--103.

\bibitem{Ng2000}
A.~Ng, S.~Russell, {Algorithms for inverse reinforcement learning}, Proceedings
  of the Seventeenth International Conference on Machine Learning 0 (2000)
  663--670.

\bibitem{NGSIM}
Next generation simulation {(NGSIM)}, \url{http://ops.fhwa.dot.gov/
  trafficanalysistools/ngsim.htm}.

\bibitem{Puterman1994}
M.~L. Puterman, Markov Decision Processes: Discrete Stochastic Dynamic
  Programming, 1st Edition, John Wiley \& Sons, Inc., New York, NY, USA, 1994.

\bibitem{Coates2009}
A.~Coates, P.~Abbeel, A.~Y. Ng, Apprenticeship learning for helicopter control,
  Communications of the ACM 52~(7) (2009) 97--105.

\bibitem{argall2009survey}
B.~D. Argall, S.~Chernova, M.~Veloso, B.~Browning, A survey of robot learning
  from demonstration, Robot. Auton. Syst. 57~(5) (2009) 469--483.

\bibitem{Boyd94}
S.~P. Boyd, L.~El~Ghaoui, E.~Feron, V.~Balakrishnan, Linear matrix inequalities
  in system and control theory, SIAM 37~(3) (1995) 479--481.

\bibitem{Baker2009_action}
C.~L. Baker, R.~Saxe, J.~B. Tenenbaum, Action understanding as inverse
  planning, Cognition 113~(3) (2009) 329 -- 349, reinforcement learning and
  higher cognition.

\bibitem{Ullman2009}
T.~D. Ullman, C.~L. Baker, O.~Macindoe, O.~Evans, N.~D. Goodman, J.~B.
  Tenenbaum, Help or hinder: Bayesian models of social goal inference, in: 22nd
  International Conference on Neural Information Processing Systems, 2009, pp.
  1874--1882.

\bibitem{Abbeel2007}
P.~Abbeel, A.~Coates, M.~Quigley, A.~Y. Ng, An application of reinforcement
  learning to aerobatic helicopter flight, in: Proceedings of the 19th
  International Conference on Neural Information Processing Systems, NIPS'06,
  MIT Press, Cambridge, MA, USA, 2006, pp. 1--8.

\bibitem{Kretzschmar_interactingpedestrians}
H.~Kretzschmar, M.~Spies, C.~Sprunk, W.~Burgard, {Socially compliant mobile
  robot navigation via inverse reinforcement learning}, The International
  Journal of Robotics Research 35~(11) (2016) 1289--1307.
\newblock \href {http://dx.doi.org/10.1177/0278364915619772}
  {\path{doi:10.1177/0278364915619772}}.

\bibitem{Kim2016_adaptivenavigation}
B.~Kim, J.~Pineau, Socially adaptive path planning in human environments using
  inverse reinforcement learning, International Journal of Social Robotics
  8~(1) (2016) 51--66.

\bibitem{Neu2007}
G.~Neu, C.~Szepesv{\'{a}}ri, {Apprenticeship Learning using Inverse
  Reinforcement Learning and Gradient Methods}, Twenty-Third Conference on
  Uncertainty in Artificial Intelligence (2007) 295--302\href
  {http://arxiv.org/abs/1206.5264} {\path{arXiv:1206.5264}}.

\bibitem{Kuderer15:Learning}
M.~Kuderer, S.~Gulati, W.~Burgard, Learning driving styles for autonomous
  vehicles from demonstration, in: IEEE International Conference on Robotics
  and Automation (ICRA), 2015, pp. 2641--2646.

\bibitem{Ziebart2008}
B.~D. Ziebart, A.~Maas, J.~A. Bagnell, A.~K. Dey, Maximum entropy inverse
  reinforcement learning, in: Proceedings of the 23rd National Conference on
  Artificial Intelligence - Volume 3, AAAI'08, AAAI Press, 2008, pp.
  1433--1438.

\bibitem{Ziebart_Cabbie}
B.~D. Ziebart, A.~L. Maas, A.~K. Dey, J.~A. Bagnell, Navigate like a cabbie:
  Probabilistic reasoning from observed context-aware behavior, in: Proceedings
  of the 10th International Conference on Ubiquitous Computing, UbiComp '08,
  ACM, New York, NY, USA, 2008, pp. 322--331.

\bibitem{Ratliff2009}
N.~D. Ratliff, D.~Silver, J.~A. Bagnell, Learning to search: Functional
  gradient techniques for imitation learning, Auton. Robots 27~(1) (2009)
  25--53.

\bibitem{Ziebart_predictionpedestrian}
B.~D. Ziebart, N.~Ratliff, G.~Gallagher, C.~Mertz, K.~Peterson, J.~A. Bagnell,
  M.~Hebert, A.~K. Dey, S.~Srinivasa, Planning-based prediction for
  pedestrians, in: Proceedings of the 2009 IEEE/RSJ International Conference on
  Intelligent Robots and Systems, IROS'09, IEEE Press, Piscataway, NJ, USA,
  2009, pp. 3931--3936.

\bibitem{Vogel_efficientdriving}
A.~Vogel, D.~Ramachandran, R.~Gupta, A.~Raux, Improving hybrid vehicle fuel
  efficiency using inverse reinforcement learning, in: AAAI Conference on
  Artificial Intelligence, 2012.

\bibitem{Bogert_mIRL_Int_2014}
K.~Bogert, P.~Doshi, Multi-robot inverse reinforcement learning under occlusion
  with state transition estimation, in: Proceedings of the 2015 International
  Conference on Autonomous Agents and Multiagent Systems, AAMAS '15,
  International Foundation for Autonomous Agents and Multiagent Systems,
  Richland, SC, 2015, pp. 1837--1838.

\bibitem{Kaelbling1996}
L.~P. Kaelbling, M.~L. Littman, A.~W. Moore, Reinforcement learning: A survey,
  J. Artif. Int. Res. 4~(1) (1996) 237--285.

\bibitem{Russell03:Artificial}
S.~Russell, P.~Norvig, Artificial Intelligence: A Modern Approach (Second
  Edition), Prentice Hall, 2003.

\bibitem{Choi2011}
J.~Choi, K.-E. Kim, Inverse reinforcement learning in partially observable
  environments, J. Mach. Learn. Res. 12 (2011) 691--730.

\bibitem{Neu2008}
G.~Neu, C.~Szepesv\'{a}ri, Training parsers by inverse reinforcement learning,
  Mach. Learn. 77~(2-3) (2009) 303--337.

\bibitem{Ratliff2006}
N.~D. Ratliff, J.~A. Bagnell, M.~A. Zinkevich, Maximum margin planning, in:
  Proceedings of the 23rd International Conference on Machine Learning, ICML
  '06, ACM, New York, NY, USA, 2006, pp. 729--736.

\bibitem{Silver2008}
A.~S. David~Silver, James~Bagnell, High performance outdoor navigation from
  overhead data using imitation learning, in: Robotics: Science and Systems IV,
  Zurich, Switzerland, 2008.

\bibitem{Abbeel2004}
P.~Abbeel, A.~Y. Ng, Apprenticeship learning via inverse reinforcement
  learning, in: Proceedings of the Twenty-first International Conference on
  Machine Learning, ICML '04, ACM, New York, NY, USA, 2004, pp. 1--8.

\bibitem{Syed2008}
U.~Syed, R.~E. Schapire, A game-theoretic approach to apprenticeship learning,
  in: Proceedings of the 20th International Conference on Neural Information
  Processing Systems, NIPS'07, Curran Associates Inc., USA, 2007, pp.
  1449--1456.

\bibitem{Jaynes_MaxEnt}
E.~T. Jaynes, Information theory and statistical mechanics, Phys. Rev. 106
  (1957) 620--630.

\bibitem{Wulfmeier2015}
M.~Wulfmeier, I.~Posner, {Maximum Entropy Deep Inverse Reinforcement Learning},
  arXiv preprint.

\bibitem{Aghasadeghi_continousspace}
N.~Aghasadeghi, T.~Bretl, Maximum entropy inverse reinforcement learning in
  continuous state spaces with path integrals, in: 2011 IEEE/RSJ International
  Conference on Intelligent Robots and Systems, 2011, pp. 1561--1566.

\bibitem{Theodorou_pathintegral}
E.~Theodorou, J.~Buchli, S.~Schaal, A generalized path integral control
  approach to reinforcement learning, J. Mach. Learn. Res. 11 (2010)
  3137--3181.

\bibitem{Boularias2012}
A.~Boularias, O.~Kr{\"o}mer, J.~Peters, Structured apprenticeship learning, in:
  P.~A. Flach, T.~De~Bie, N.~Cristianini (Eds.), Machine Learning and Knowledge
  Discovery in Databases: European Conference, ECML PKDD 2012, Bristol, UK,
  September 24-28, 2012. Proceedings, Part II, Springer Berlin Heidelberg,
  Berlin, Heidelberg, 2012, pp. 227--242.

\bibitem{Kullback68informationtheory}
S.~Kullback, Information theory and statistics (1968).

\bibitem{Boularias2011relative}
A.~Boularias, J.~Kober, J.~Peters, Relative entropy inverse reinforcement
  learning, in: Proceedings of the Fourteenth International Conference on
  Artificial Intelligence and Statistics, {AISTATS} 2011, Fort Lauderdale, USA,
  April 11-13, 2011, 2011, pp. 182--189.

\bibitem{Ramachandran2007}
D.~Ramachandran, E.~Amir, Bayesian inverse reinforcement learning, in:
  Proceedings of the 20th International Joint Conference on Artifical
  Intelligence, IJCAI'07, Morgan Kaufmann Publishers Inc., San Francisco, CA,
  USA, 2007, pp. 2586--2591.

\bibitem{Choi11:MAP}
J.~Choi, K.~eung Kim, Map inference for bayesian inverse reinforcement
  learning, in: Advances in Neural Information Processing Systems 24, 2011, pp.
  1989--1997.

\bibitem{Lopes2009}
M.~Lopes, F.~Melo, L.~Montesano, Active learning for reward estimation in
  inverse reinforcement learning, in: Proceedings of the European Conference on
  Machine Learning and Knowledge Discovery in Databases: Part II, ECML PKDD
  '09, Springer-Verlag, Berlin, Heidelberg, 2009, pp. 31--46.

\bibitem{Levine2011}
S.~Levine, Z.~Popovi\'{c}, V.~Koltun, Nonlinear inverse reinforcement learning
  with gaussian processes, in: Proceedings of the 24th International Conference
  on Neural Information Processing Systems, NIPS'11, Curran Associates Inc.,
  USA, 2011, pp. 19--27.

\bibitem{Babes-Vroman2011}
M.~Babes-Vroman, V.~Marivate, K.~Subramanian, M.~Littman, Apprenticeship
  learning about multiple intentions, in: 28th International Conference on
  Machine Learning, ICML 2011, 2011, pp. 897--904.

\bibitem{Klein2012structured}
E.~Klein, M.~Geist, B.~Piot, O.~Pietquin, Inverse reinforcement learning
  through structured classification, in: Proceedings of the 25th International
  Conference on Neural Information Processing Systems, NIPS'12, Curran
  Associates Inc., USA, 2012, pp. 1007--1015.

\bibitem{taskar2005learning}
B.~Taskar, V.~Chatalbashev, D.~Koller, C.~Guestrin, Learning structured
  prediction models: A large margin approach, in: 22nd International Conference
  on Machine Learning, 2005, p. 896–903.

\bibitem{Klein_CSI}
E.~Klein, B.~Piot, M.~Geist, O.~Pietquin, A cascaded supervised learning
  approach to inverse reinforcement learning, in: Proceedings of the European
  Conference on Machine Learning and Knowledge Discovery in Databases - Volume
  8188, ECML PKDD 2013, Springer-Verlag New York, Inc., New York, NY, USA,
  2013, pp. 1--16.

\bibitem{brown19a}
D.~Brown, W.~Goo, P.~Nagarajan, S.~Niekum, Extrapolating beyond suboptimal
  demonstrations via inverse reinforcement learning from observations, in:
  Proceedings of the 36th International Conference on Machine Learning, Vol.~97
  of Proceedings of Machine Learning Research, 2019, pp. 783--792.

\bibitem{Levine2010_featureconstruction}
S.~Levine, Z.~Popovi\'{c}, V.~Koltun, Feature construction for inverse
  reinforcement learning, in: Proceedings of the 23rd International Conference
  on Neural Information Processing Systems, NIPS'10, Curran Associates Inc.,
  USA, 2010, pp. 1342--1350.

\bibitem{ZiebartBD_compare_MMP}
B.~D. Ziebart, J.~A. Bagnell, A.~K. Dey, Modeling interaction via the principle
  of maximum causal entropy, in: J.~Fürnkranz, T.~Joachims (Eds.), Proceedings
  of the 27th International Conference on Machine Learning (ICML-10),
  Omnipress, 2010, pp. 1255--1262.

\bibitem{brown2018efficient}
D.~S. Brown, S.~Niekum, Efficient probabilistic performance bounds for inverse
  reinforcement learning, in: Thirty-Second AAAI Conference on Artificial
  Intelligence, 2018.

\bibitem{Coates2008}
A.~Coates, P.~Abbeel, A.~Y. Ng, Learning for control from multiple
  demonstrations, in: Proceedings of the 25th International Conference on
  Machine Learning, ICML '08, ACM, New York, NY, USA, 2008, pp. 144--151.

\bibitem{Melo_perturbedinput}
F.~S. Melo, M.~Lopes, R.~Ferreira, Analysis of inverse reinforcement learning
  with perturbed demonstrations, in: Proceedings of the 2010 Conference on ECAI
  2010: 19th European Conference on Artificial Intelligence, IOS Press,
  Amsterdam, The Netherlands, The Netherlands, 2010, pp. 349--354.

\bibitem{Shiarlis_2016_failed}
K.~Shiarlis, J.~Messias, S.~Whiteson, Inverse reinforcement learning from
  failure, in: Proceedings of the 2016 International Conference on Autonomous
  Agents and Multiagent Systems, AAMAS '16, International Foundation for
  Autonomous Agents and Multiagent Systems, Richland, SC, 2016, pp. 1060--1068.

\bibitem{Ziebart2010}
B.~Ziebart, {Modeling Purposeful Adaptive Behavior with the Principle of
  Maximum Causal Entropy}, Ph.D. thesis, Carnegie Mellon University (December
  2010).

\bibitem{Grunwald2004}
P.~D. Gr{\"{u}}nwald, A.~P. Dawid, {Game theory, maximum entropy, minimum
  discrepancy and robust bayesian decision theory}, The Annals of Statistics
  32~(1) (2004) 1367--1433.
\newblock \href {http://arxiv.org/abs/0410076v1} {\path{arXiv:0410076v1}}.

\bibitem{Dimitrakakis2012}
C.~Dimitrakakis, C.~A. Rothkopf, Bayesian multitask inverse reinforcement
  learning, in: Proceedings of the 9th European Conference on Recent Advances
  in Reinforcement Learning, EWRL'11, Springer-Verlag, Berlin, Heidelberg,
  2012, pp. 273--284.

\bibitem{Syed2008_supplement}
U.~Syed, R.~E. Schapire, {A Game-Theoretic Approach to Apprenticeship
  Learning—Supplement} (2007).

\bibitem{Vroman2014}
M.~C. Vroman, {MAXIMUM LIKELIHOOD INVERSE REINFORCEMENT LEARNING}, Ph.D.
  thesis, Rutgers, The State University of New Jersey (2014).

\bibitem{Lee_Improved_Projection}
S.~J. Lee, Z.~Popovi\'{c}, Learning behavior styles with inverse reinforcement
  learning, ACM Trans. Graph. 29~(4) (2010) 122:1--122:7.

\bibitem{Finn_gcl}
C.~Finn, S.~Levine, P.~Abbeel, {Guided Cost Learning: Deep Inverse Optimal
  Control via Policy Optimization}, arXiv preprint arXiv:1603.00448.

\bibitem{S.Melo2010}
F.~S. Melo, M.~Lopes, Learning from demonstration using mdp induced metrics,
  in: Proceedings of the 2010 European Conference on Machine Learning and
  Knowledge Discovery in Databases: Part II, ECML PKDD'10, Springer-Verlag,
  Berlin, Heidelberg, 2010, pp. 385--401.

\bibitem{Munzer_relationalIRL}
T.~Munzer, B.~Piot, M.~Geist, O.~Pietquin, M.~Lopes, Inverse reinforcement
  learning in relational domains, in: Proceedings of the 24th International
  Conference on Artificial Intelligence, IJCAI'15, AAAI Press, 2015, pp.
  3735--3741.

\bibitem{fletcher1987practical}
R.~Fletcher, Practical methods of optimization, Wiley-Interscience publication,
  Wiley, 1987.

\bibitem{Malouf_ComparisonEstimatns}
R.~Malouf, A comparison of algorithms for maximum entropy parameter estimation,
  in: Proceedings of the 6th Conference on Natural Language Learning - Volume
  20, COLING-02, Association for Computational Linguistics, Stroudsburg, PA,
  USA, 2002, pp. 1--7.

\bibitem{Vernaza_highdimension_approximation}
P.~Vernaza, J.~A. Bagnell, Efficient high-dimensional maximum entropy modeling
  via symmetric partition functions, in: Proceedings of the 25th International
  Conference on Neural Information Processing Systems, NIPS'12, Curran
  Associates Inc., USA, 2012, pp. 575--583.

\bibitem{Kolter_2007_HierarchicalAL}
J.~Z. Kolter, P.~Abbeel, A.~Y. Ng, Hierarchical apprenticeship learning, with
  application to quadruped locomotion, in: Proceedings of the 20th
  International Conference on Neural Information Processing Systems, NIPS'07,
  Curran Associates Inc., USA, 2007, pp. 769--776.

\bibitem{Rothkopf_modular_highDim}
C.~A. Rothkopf, D.~H. Ballard, Modular inverse reinforcement learning for
  visuomotor behavior, Biol. Cybern. 107~(4) (2013) 477--490.

\bibitem{Wang2002}
S.~Wang, R.~Rosenfeld, Y.~Zhao, D.~Schuurmans, {The Latent Maximum Entropy
  Principle}, in: IEEE International Symposium on Information Theory, 2002, pp.
  131--131.

\bibitem{Wang2012}
S.~Wang, D.~{Schuurmans Yunxin Zhao}, {The Latent Maximum Entropy Principle},
  ACM Transactions on Knowledge Discovery from Data 6~(8).

\bibitem{Bogert_EM_hiddendata_fruit}
K.~Bogert, J.~F.-S. Lin, P.~Doshi, D.~Kulic, Expectation-maximization for
  inverse reinforcement learning with hidden data, in: Proceedings of the 2016
  International Conference on Autonomous Agents and Multiagent Systems, AAMAS
  '16, International Foundation for Autonomous Agents and Multiagent Systems,
  2016, pp. 1034--1042.

\bibitem{hiddenMDP_Kitani2012}
K.~M. Kitani, B.~D. Ziebart, J.~A. Bagnell, M.~Hebert, Activity forecasting,
  in: Proceedings of the 12th European Conference on Computer Vision - Volume
  Part IV, ECCV'12, Springer-Verlag, Berlin, Heidelberg, 2012, pp. 201--214.

\bibitem{Kaelbling_pomdp}
L.~P. Kaelbling, M.~L. Littman, A.~R. Cassandra, Planning and acting in
  partially observable stochastic domains, Artif. Intell. 101~(1-2) (1998)
  99--134.

\bibitem{Choi2012}
J.~Choi, K.-E. Kim, Nonparametric bayesian inverse reinforcement learning for
  multiple reward functions, in: Proceedings of the 25th International
  Conference on Neural Information Processing Systems, NIPS'12, Curran
  Associates Inc., USA, 2012, pp. 305--313.

\bibitem{Reddy_DecMulti}
T.~S. Reddy, V.~Gopikrishna, G.~Zaruba, M.~Huber, Inverse reinforcement
  learning for decentralized non-cooperative multiagent systems, in: 2012 IEEE
  International Conference on Systems, Man, and Cybernetics (SMC), 2012, pp.
  1930--1935.

\bibitem{Lin_MultiGame}
X.~Lin, P.~A. Beling, R.~Cogill, Multi-agent inverse reinforcement learning for
  zero-sum games, CoRR abs/1403.6508.

\bibitem{Bogert_mIRL_woT_Int_2015}
K.~Bogert, P.~Doshi, Toward estimating others' transition models under
  occlusion for multi-robot irl, in: Proceedings of the 24th International
  Conference on Artificial Intelligence, IJCAI'15, AAAI Press, 2015, pp.
  1867--1873.

\bibitem{Levine_UnknownDynamics_NN_policy}
S.~Levine, P.~Abbeel, Learning neural network policies with guided policy
  search under unknown dynamics, in: Proceedings of the 27th International
  Conference on Neural Information Processing Systems, NIPS'14, MIT Press,
  Cambridge, MA, USA, 2014, pp. 1071--1079.

\bibitem{Ratliff2007}
N.~Ratliff, D.~Bradley, J.~A. Bagnell, J.~Chestnutt, Boosting structured
  prediction for imitation learning, in: Proceedings of the 19th International
  Conference on Neural Information Processing Systems, NIPS'06, MIT Press,
  Cambridge, MA, USA, 2006, pp. 1153--1160.

\bibitem{Ng_RegularizationComparison}
A.~Y. Ng, Feature selection, l1 vs. l2 regularization, and rotational
  invariance, in: Proceedings of the Twenty-first International Conference on
  Machine Learning, ICML '04, ACM, New York, NY, USA, 2004, pp. 78--.

\bibitem{Choi2013bayesian}
J.~Choi, K.-E. Kim, Bayesian nonparametric feature construction for inverse
  reinforcement learning, in: Proceedings of the Twenty-Third International
  Joint Conference on Artificial Intelligence, IJCAI '13, AAAI Press, 2013, pp.
  1287--1293.

\bibitem{yang2009cuckoo}
X.-S. Yang, S.~Deb, Cuckoo search via l{\'e}vy flights, in: 2009 World Congress
  on Nature \& Biologically Inspired Computing (NaBIC), IEEE, 2009, pp.
  210--214.

\bibitem{yang2010engineering}
X.-S. Yang, S.~Deb, Engineering optimisation by cuckoo search, arXiv preprint
  arXiv:1005.2908.

\bibitem{eberhart1995particle}
R.~Eberhart, J.~Kennedy, Particle swarm optimization, in: Proceedings of the
  IEEE international conference on neural networks, Vol.~4, Citeseer, 1995, pp.
  1942--1948.

\bibitem{yang2010firefly}
X.-S. Yang, Firefly algorithm, stochastic test functions and design
  optimisation, arXiv preprint arXiv:1003.1409.

\bibitem{Littman1994markov_games}
M.~L. Littman, Markov games as a framework for multi-agent reinforcement
  learning, in: Proceedings of the eleventh international conference on machine
  learning, Vol. 157, 1994, pp. 157--163.

\bibitem{Boutilier1999}
C.~Boutilier, Sequential optimality and coordination in multiagent systems, in:
  Proceedings of the 16th International Joint Conference on Artifical
  Intelligence - Volume 1, IJCAI'99, Morgan Kaufmann Publishers Inc., San
  Francisco, CA, USA, 1999, pp. 478--485.

\bibitem{Spaan_InteractiveFulObs}
M.~T.~J. Spaan, F.~S. Melo, Interaction-driven markov games for decentralized
  multiagent planning under uncertainty, in: Proceedings of the 7th
  International Joint Conference on Autonomous Agents and Multiagent Systems -
  Volume 1, AAMAS '08, International Foundation for Autonomous Agents and
  Multiagent Systems, Richland, SC, 2008, pp. 525--532.

\bibitem{Peshkin00}
L.~Peshkin, K.-E. Kim, N.~Meuleau, L.~P. Kaelbling, Learning to cooperate via
  policy search, in: Proceedings of the 16th Conference on Uncertainty in
  Artificial Intelligence, UAI '00, Morgan Kaufmann Publishers Inc., San
  Francisco, CA, USA, 2000, pp. 489--496.

\bibitem{PynadathT02}
D.~V. Pynadath, M.~Tambe, The communicative multiagent team decision problem:
  Analyzing teamwork theories and models, J. Artif. Int. Res. 16~(1) (2002)
  389--423.

\bibitem{Bernstein02}
D.~S. Bernstein, R.~Givan, N.~Immerman, S.~Zilberstein, The complexity of
  decentralized control of markov decision processes, Math. Oper. Res. 27~(4)
  (2002) 819--840.

\bibitem{Gmytrasiewicz_Doshi_IPOMDP}
P.~J. Gmytrasiewicz, P.~Doshi, A framework for sequential planning in
  multi-agent settings, J. Artif. Int. Res. 24~(1) (2005) 49--79.

\bibitem{WaughZB_InverseEquilibrium}
K.~Waugh, B.~D. Ziebart, J.~A. Bagnell, Computational rationalization: The
  inverse equilibrium problem, CoRR abs/1308.3506.

\bibitem{Kalakrishnan_continousspace}
M.~Kalakrishnan, P.~Pastor, L.~Righetti, S.~Schaal, Learning objective
  functions for manipulation, in: IEEE International Conference on Robotics and
  Automation (ICRA), 2013, 2013, pp. 1331--1336.

\end{thebibliography}

\end{document}